\documentclass[journal]{IEEEtran}
\usepackage[T1]{fontenc}
\usepackage{amsmath,amsfonts}

\usepackage[cmintegrals]{newtxmath}

\usepackage{setspace}

\usepackage[ruled]{algorithm2e} 
\usepackage{array}
\usepackage{tabularx}
\usepackage{textcomp}
\usepackage{stfloats}
\usepackage{url}
\usepackage{verbatim}
\usepackage{graphicx}
\usepackage{cite}
\usepackage{color}
\usepackage{float} 
\usepackage{subfigure}

\usepackage{amssymb}
\usepackage{booktabs}
\usepackage{multirow}
\usepackage[table,xcdraw]{xcolor}
\usepackage{hyperref}

\usepackage{tikz}

\usepackage[commandnameprefix=always]{changes}

\begin{document}

\title{Bridging Digital Deception and Physical Fidelity: A Spatio-Temporal Attack on Cognitive Radars}

 \author{Tongrun Xing, Yongjun Dai, Xuan Liao, Zhiwen Zhang, Tong Zhou, Longlong Zhang and Tian Jiang

 \thanks{This work was supported by the Innovation Research Foundation of National University of Defense Technology (ZK25-06). (Corresponding authors: Longlong Zhang and Tian Jiang.)}
 \thanks{Tongrun Xing, Yongjun Dai, Xuan Liao and Zhiwen Zhang are with the College of Computer Science and Technology, National University of Defense Technology, Changsha 410073, China. (e-mail: xingtongrun01@nudt.edu.cn, daiyongjun23@nudt.edu.cn, liaoxuan21@nudt.edu.cn,  zhangzhiwen20@nudt.edu.cn)}
  \thanks{Tong Zhou is with the College of Advanced Interdisciplinary Studies, National University of Defense Technology, Changsha 410073, China. (e-mail: zhoutong09@nudt.edu.cn)}
  \thanks{Longlong Zhang is with the College of Science, National University of Defense Technology, Changsha 410073, China. (e-mail: zll@nudt.edu.cn)}
 \thanks{Tian Jiang is with the College of Science, National University of Defense Technology, Changsha 410073,
China, also with the Hunan Research Center of the Basic Discipline for Physical States, National University of Defense Technology,  Changsha 410073, China. (e-mail: tjiang@nudt.edu.cn)}}

\markboth{Preprint}{Xing \MakeLowercase{\textit{et al.}}: Segmented Spatio-Temporal Attack}

\maketitle
\begin{abstract}
As cognitive radars evolve towards the end-to-end processing of one-dimensional complex baseband sequences, deep networks face severe adversarial threats. However, existing attacks encounter dual limitations in practical electronic warfare: low cross-architecture transferability and severe vulnerability to physical channel distortions, which degrades their tactical reliability. To overcome these bottlenecks, we propose a physical prior-driven Segmented Spatio-Temporal Attack (SSTA). First, a global-local feature decoupling segmentation mechanism interrupts the adversarial gradients' dependence on long-range sequence coherence, bridging the cross-architecture feature gap. Second, moving beyond static norm constraints devoid of physical semantics, we abstract channel distortions into differentiable spatio-temporal operators, integrating them into momentum iterations via an Expectation over Transformation (EoT) strategy. Comprehensive cross-evaluations across 9 heterogeneous models spanning CNNs, RNNs, and Transformers demonstrate that SSTA outperforms the state-of-the-art baseline by 10.36\% in Average Black-Box Attack Success Rate. Furthermore, baseband equivalent simulations confirm that the generated waveforms survive extreme physical distortions and, after matched filtering, achieve zero main-lobe attenuation ($0\ \mathrm{dB}$) with noise floor lift strictly within Constant False Alarm Rate tolerances. SSTA successfully unifies digital-domain deception and physical-domain fidelity, providing crucial support for CEW security assessment.
\end{abstract}



\section{Introduction} \label{sec1}

\IEEEPARstart{W}{ith} the rapid development of digital array radar and Cognitive Electronic Warfare (CEW), deep learning technologies have been widely applied in the field of radar active jamming recognition \cite{Zan2025, Yu2023, Ref_FSLVGCN, Wu2023, Wang2023}. Currently, end-to-end cognitive networks utilizing one-dimensional (1D) complex baseband sequences (I/Q dual-channel) as direct inputs are gradually becoming the mainstream development direction, as they can effectively preserve intrinsic electromagnetic characteristics. However, while enhancing recognition accuracy, this input mechanism directly exposes deep cognitive models to the attack surface of minute electromagnetic perturbations, triggering security concerns regarding Adversarial Examples \cite{9766209, 11440110, Jiahao2025}. Unlike in the computer vision domain, signal detection in real radar systems highly relies on the physical radio frequency (RF) front-end and strict coherent processing chains (e.g., matched filtering) \cite{richards2014fundamentals}. In electromagnetic engagements, if minute adversarial perturbations can be covertly superimposed onto the original jamming echoes and maintain their perturbation features after processing by high-gain radar receivers, they can induce high-confidence misclassifications by the cognitive network, subsequently leading to target misidentification or track loss \cite{2019Adversarial}. This adversarial threat, operating at the physical layer of 1D RF signals and requiring penetration of physical processing chains, constitutes a critical security issue that urgently needs to be addressed in current CEW.

However, existing adversarial attack algorithms encounter two core challenges when faced with practical radar electronic warfare scenarios. The first is the feature transfer barrier across heterogeneous architectures. Under strict black-box conditions, attackers cannot access the internal parameters of the adversary's radar models \cite{Papernot2016Transferability, Liu2017Delving, XiaosenLuo2026Devling}. To mitigate overfitting to a single white-box model, existing state-of-the-art black-box attack algorithms have made extensive explorations. Specifically, these advanced strategies can be broadly categorized into gradient optimization methods based on momentum evolution \cite{Dong2018Boosting, Jiadong2020NESTEROV, Xiaosen2021}, input transformation methods aimed at feature disruption \cite{Xie2018Improving, Yinpeng2019Evading, Wang2023Structure, WangKun2021Admix}, and deep representation tuning strategies \cite{AdityaGaneshan2019FDA, Huang2019Enhancing, Wang2021Feature, Wang2024Improving}. Nevertheless, the underlying logic of these algorithms is primarily based on the visual priors of 2D images (such as translation invariance and spatial redundancy). When directly applied to 1D radar sequences with strict temporal dependencies and phase coupling, adversarial perturbations are highly prone to overfitting the source model's local receptive fields and specific absolute positional encodings. Consequently, when transferring across heterogeneous defense networks (e.g., from local convolution-based CNNs to global self-attention-based Transformers \cite{Luo2023}), the attack success rate decreases significantly. The second challenge is the lack of tactical equivalence under real physical RF constraints. Traditional adversarial perturbations are mostly generated in a static digital domain. Once they encounter clock synchronization drift, orthogonal phase rotation, or noise floor fluctuations of non-cooperative receivers in actual electromagnetic channels, their adversarial features are vulnerable to fading and thus become ineffective. Furthermore, some communication or speech adversarial methods relying on auditory masking effects or global static norm constraints \cite{Nicholas2018Audio, Xuejing2018CommanderSong} struggle to maintain waveform stealthiness under the stringent pulse compression mechanisms of radars. These limitations in channel distortion resistance and coherent processing fidelity make it difficult for existing algorithms to maintain tactical equivalence in practical jamming scenarios.

To effectively bridge the cross-architecture feature gap and satisfy the physical constraints of radar baseband receiving chains, this paper departs from the traditional static optimization framework in the purely digital domain and proposes a physical prior-driven Segmented Spatio-Temporal Attack (SSTA). This method operates directly on 1D radar complex baseband sequences, aiming to enhance the black-box transfer rate while effectively accommodating the physical characteristics of radar coherent processing.

The main contributions of this paper can be summarized in the following three aspects:

\begin{enumerate}
    \item A segmentation mechanism based on global-local feature decoupling is proposed to mitigate the cross-architecture black-box transfer barrier. By independently slicing and locally optimizing the baseband sequence along the time axis, this mechanism effectively blocks the dependence of adversarial gradients on the long-range coherence of a single surrogate network. This strategy of extracting generalized features within discrete local spaces reduces the feature aggregation discrepancy between heterogeneous cognitive architectures, thereby significantly improving the cross-model black-box transfer success rate.
    
    \item A physical prior-driven spatio-temporal Expectation over Transformation (EoT) strategy is constructed to accommodate both model generalization and physical receiving chain compatibility. This paper abstracts channel distortions (clock drift, Jamming-to-Noise Ratio (JNR) fluctuations, and orthogonal phase rotations) into differentiable spatio-temporal joint operators and incorporates them into the momentum iteration trajectory of gradients. This strategy not only achieves smooth regularization of the loss landscape, further enhancing the generalization capability of the algorithm, but also endows the generated adversarial waveforms with the physical robustness to resist channel fading and adapt to matched filtering.
    
    \item Comprehensive cross-evaluations across multiple heterogeneous models and equivalent simulations of radar baseband are conducted, achieving an effective integration of digital-domain deception and physical-domain fidelity. In 72 cross-evaluation paths comprising 9 heterogeneous models spanning CNNs, RNNs, and Transformers, SSTA achieves an increase of over 10\% improvement in Average Black-Box Attack Success Rate compared to existing baseline algorithms. In addition, equivalent simulations based on radar physical receiving chains confirm that the generated adversarial waveforms can maintain a high effective survival rate under temporal, phase, and energy distortions. After passing through the radar matched filter, they cause no obvious attenuation of the main-lobe energy, and the noise floor lift is controlled within the Constant False Alarm Rate (CFAR) tolerance, demonstrating favorable application potential.
\end{enumerate}

The remainder of this paper is organized as follows: Section \ref{sec2} defines the radar baseband signals, physical distortion models, and the adversarial optimization objective with expectation over transformation. Section \ref{sec3} details the derivation of the SSTA framework, explaining the mathematical principles of the feature decoupling segmentation mechanism and the physical spatio-temporal operators. Section \ref{sec4} conducts cross-generalization evaluations based on a matrix of 9 heterogeneous models and verifies the tactical equivalence and physical robustness of the generated waveforms. Finally, Section \ref{sec5} concludes the paper and discusses future perspectives.

\section{System Model and Problem Formulation} \label{sec2}

\subsection{1D Baseband Signal and Coherent Deception Jamming Model}

The receiving end of modern digital array radars maps the intercepted radio frequency electromagnetic waves into one-dimensional complex baseband sequences through quadrature demodulation and analog-to-digital conversion. In this paper, the radar baseband signal is rigorously defined as a complex vector $\mathbf{x} \in \mathbb{C}^{1 \times L}$ of length $L$, which comprises In-phase (I) and Quadrature (Q) dual-channel data, denoted as $\mathbf{x} = \mathbf{x}_{\mathrm{I}} + \mathrm{j}\mathbf{x}_{\mathrm{Q}}$.

The essence of active coherent deception jamming is to intercept, modulate, and retransmit the radar's transmitted signal to generate realistic false targets at the receiving end \cite{adamy2004ew102, zhao2012radar}. Let the complex baseband sequence of the original radar transmitted waveform be $\mathbf{s} \in \mathbb{C}^{1 \times L}$. Typical coherent jamming (e.g., Dense False Target Jamming, DFTJ) is constructed by superimposing multiple discrete pulse segments along the time axis, each with an independent delay $\tau_i$, amplitude $A_i$, and phase shift $\phi_i$. Its discrete-time mathematical model can be expressed as:
\begin{equation}
x[n] = \sum_{i=1}^{K} A_i s[n - \tau_i] e^{\mathrm{j}\phi_i}
\end{equation}

Unlike static images in the computer vision domain, target detection in radar systems highly depends on a strict coherent processing chain, the core component of which is pulse compression (matched filtering). Suppose the impulse response of the receiver's matched filter is designed as the time-reversed and conjugated version of the transmitted signal, i.e., $h[n] = s^*[-n]$ \cite{richards2014fundamentals}. The output $y[n]$ of the received jamming sequence $\mathbf{x}$ after matched filtering manifests as a discrete convolution:
\begin{equation}
y[n] = \sum_{k} x[k] h[n-k]
\end{equation}

Because the coherent jamming signal retains the pulse intra-phase history of the original radar transmitted pulse, it can achieve coherent accumulation of energy during the convolution process, forming high-power false spikes. To quantify the tactical fidelity of adversarial perturbations within the physical processing chain, this paper introduces two core metrics: Peak Attenuation ($\Delta P_{\mathrm{peak}}$) and Noise Floor Lift ($\Delta P_{\mathrm{noise}}$) to evaluate the quality of the 1D Range Profile after pulse compression.

Let $y_{\mathrm{clean}}[n]$ and $y_{\mathrm{adv}}[n]$ denote the matched filtering output sequences of the original clean jamming and the jamming with added adversarial perturbations, respectively. Peak Attenuation is defined as the logarithmic difference of their global maximum peak powers:
\begin{equation}
\Delta P_{\mathrm{peak}} = 10 \log_{10} \frac{\max_n |y_{\mathrm{clean}}[n]|^2}{\max_n |y_{\mathrm{adv}}[n]|^2} \ (\mathrm{dB})
\end{equation}
This metric strictly quantifies the degree of energy loss inflicted by the adversarial perturbation on the core energy of the original deception jamming. Furthermore, Noise Floor Lift quantifies the extent to which the perturbation energy disperses into non-main-lobe regions after pulse compression. Let $\Omega_{\mathrm{noise}}$ be the noise floor interval in the range profile excluding the main lobe and near side lobes. This metric is defined as the growth ratio of the average power within this interval:
\begin{equation}
\Delta P_{\mathrm{noise}} = 10 \log_{10} \frac{\sum_{n \in \Omega_{\mathrm{noise}}} |y_{\mathrm{adv}}[n]|^2}{\sum_{n \in \Omega_{\mathrm{noise}}} |y_{\mathrm{clean}}[n]|^2} \ (\mathrm{dB})
\end{equation}

When launching adversarial attacks, if unconstrained digital perturbations disrupt the physical envelope coherence of the original signal, it will lead to a sharp decrease in $\Delta P_{\mathrm{peak}}$ and a drastic lift in $\Delta P_{\mathrm{noise}}$ after matched filtering. This not only significantly weakens the deceptive power of the false targets but also triggers abnormal alarms in the radar's Constant False Alarm Rate detection mechanism due to the abnormally elevated wideband noise floor, thereby exposing the jamming intention and losing tactical equivalence.

\subsection{Electromagnetic Channel Distortion Model}

Existing digital-domain adversarial attack algorithms are typically established on the assumption of ideal static data distributions, which essentially differs from real radar electronic warfare spaces. In non-cooperative electromagnetic engagements, when adversary reconnaissance receivers intercept signals and implement jamming, complex physical distortions are inevitably introduced. To endow adversarial examples with practical survivability during the generation phase, this paper abstracts three core types of channel fading, which are unavoidable in actual physical chains, into a mathematical operator set $\mathcal{T}$.

First, due to clock asynchrony in non-cooperative systems and dynamic changes in spatial distance, the baseband sequences acquired by the receiving system experience random drift along the time axis. Let $\tau \in \mathbb{Z}$ be the discrete time delay offset; the clock synchronization drift model can be expressed as a circular shift of the sequence:
\begin{equation}
T_{\mathrm{time}}(\mathbf{x})[n] = x[(n - \tau) \bmod L]
\end{equation}

Second, minute deviations in the receiver's local oscillator frequency or residual Doppler shifts caused by the relative motion of targets lead to energy coupling between the $\mathrm{I}/\mathrm{Q}$ dual channels of the baseband signal. Let $\phi \in [-\pi, \pi)$ be the introduced random phase rotation angle; the orthogonal phase rotation model can be defined as:
\begin{equation}
T_{\mathrm{phase}}(\mathbf{x})[n] = x[n] e^{\mathrm{j}\phi}
\end{equation}

Finally, the real electromagnetic space is filled with complex background clutter and receiver thermal noise. Under a given JNR, the channel noise model can be formulated as superimposing independent identically distributed (i.i.d.) complex Gaussian white noise $\mathbf{w} \sim \mathcal{CN}(\mathbf{0}, \sigma^2 \mathbf{I})$ onto the original sequence:
\begin{equation}
T_{\mathrm{noise}}(\mathbf{x}) = \mathbf{x} + \mathbf{w}
\end{equation}

Traditional adversarial perturbations rely heavily on the absolute spatiotemporal coordinates of the original input $x$. As highlighted by existing studies \cite{athalye2018synthesizing, 2019Adversarial}, once subjected to the aforementioned channel distortions, including $T_{\text{time}}$, $T_{\text{phase}}$ and $T_{\text{noise}}$, during actual over-the-air (OTA) propagation, the fragile decision boundaries constructed along the direction of minute gradients become highly susceptible to failure.

\subsection{Threat Model and Adversarial Formulation}

This paper strictly defines the adversarial scenario as a Black-box Attack. Under this threat model, the attacker possesses only the same physical signal transceiving capabilities as the adversary radar and can obtain the data structure and feature scheme of the input signals, but has no access to the internal architecture, neuron weights, and backpropagation gradients of the adversary's cognitive radar defense system $f_{\mathrm{tgt}}$. The attacker can only generate adversarial perturbations $\boldsymbol{\delta}$ based on a locally constructed white-box surrogate model $f_{\mathrm{src}}$ and rely on the generalization of its feature space to achieve cross-model transfer.

Considering the dynamic range limitations of the Digital-to-Analog Converter in the radar transmitter, and to prevent the perturbation from triggering the radar receiver's abnormal alarm mechanisms due to excessive energy, this paper adopts a strict infinity norm constraint ($L_\infty$ norm) to define the physical boundary of the perturbation. Specifically, the generated adversarial perturbation $\boldsymbol{\delta}$ must satisfy $\|\boldsymbol{\delta}\|_\infty \le \epsilon$, where $\epsilon$ is the minimal perturbation budget, ensuring that the perturbation is safely hidden within the radar system's quantization noise floor and natural clutter.

Combining the coherent physical constraints of baseband signals (Section II-A) and the electromagnetic channel distortion models (Section II-B), this paper formulates the black-box adversarial attack on 1D radar signals as a constrained optimization problem with physical EoT. Given the true label $y$ and the surrogate model's loss function $\mathcal{L}(\cdot)$, the global optimization objective of the proposed algorithm to solve for the optimal adversarial perturbation $\boldsymbol{\delta}$ is formulated as follows:
\begin{equation}
\begin{aligned}
\arg \max_{\boldsymbol{\delta}} \quad & \mathbb{E}_{T \sim \mathcal{T}} \left[ \mathcal{L}(f_{\mathrm{src}}(T(\mathbf{x} + \boldsymbol{\delta})), y) \right] \\
\text{s.t.} \quad & \|\boldsymbol{\delta}\|_\infty \le \epsilon
\end{aligned}
\end{equation}

where $T \sim \mathcal{T}$ represents the sampling operation from the true physical distortion distribution jointly composed of time delays, phase shifts, and channel noise. This objective function explicitly establishes from a mathematical foundation that: effective adversarial perturbations should not solely pursue loss maximization on a static input $\mathbf{x}$, but must maintain their destructive attack capability even after experiencing various potential physical fading. This provides a rigorous theoretical basis for deriving the spatio-temporal joint operators and momentum iteration strategies based on physical priors within the SSTA framework in Section \ref{sec3}.

\section{Segmented Spatio-Temporal Attack (SSTA)} \label{sec3}

\subsection{Overall Framework of SSTA}

\begin{figure*}[htbp]
    \centering
    \includegraphics[width=0.98\textwidth]{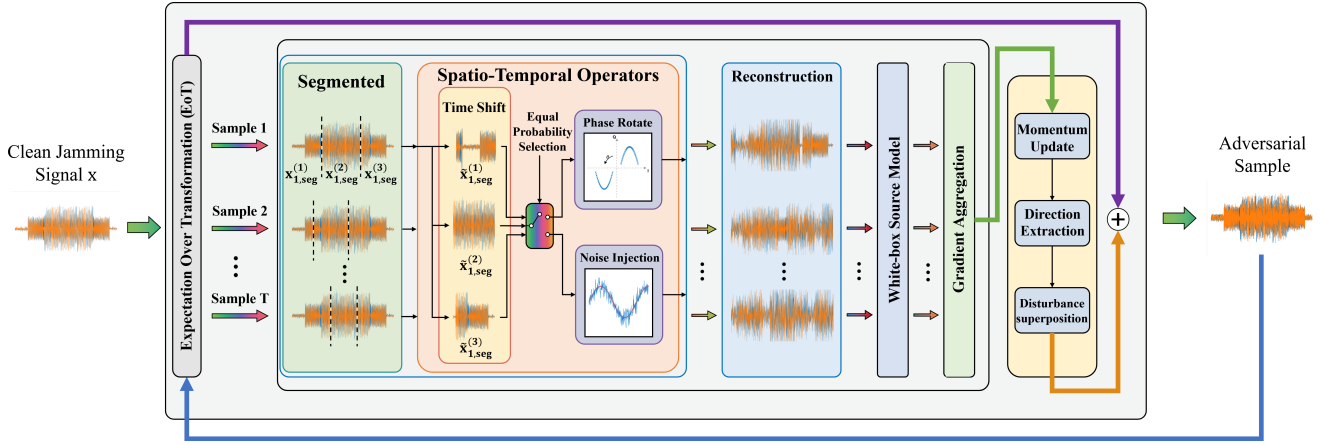}
    \caption{The overall framework of Segmented Spatio-Temporal Attack.}
    \label{fig:framework}
\end{figure*}

To overcome the cross-architecture feature barriers and physical chain disconnections analyzed in Section \ref{sec2}, this paper steps outside the purely data-driven optimization framework and proposes the SSTA. The core motivation of SSTA lies in deeply integrating the physical priors inherent in electromagnetic channels and radar hardware systems into the generation process of adversarial perturbations. 

The overall operational mechanism of the SSTA algorithm is illustrated in Fig. \ref{fig:framework}, and its entire architecture consists of three mutually synergistic core mechanisms:
1) A global-local feature decoupling segmentation mechanism that slices the original radar signal to forcefully interrupt the continuous propagation of adversarial gradients along the time axis.
2) Physical prior-driven spatio-temporal dual-domain operators that instantiate the electromagnetic distortions (synchronization delays, power fluctuations, and phase rotations) as differentiable forward operators.
3) A tactical-equivalent EoT cascading strategy that performs asymmetric random sampling in each iteration, confining the perturbation energy to the low-sensitivity regions of radar matched filtering. 

These three mechanisms are deeply embedded into the momentum iterative optimization loop, ensuring that the generated adversarial examples exhibit outstanding black-box transferability while preserving the tactical effectiveness of the original deception jamming signals.

\subsection{Physical Prior-Driven Spatio-Temporal Operators}

To accommodate the input format of deep neural networks, the complex baseband signal defined in Section \ref{sec2} is mapped into a dual-channel real-valued tensor. Based on the electromagnetic channel distortion models established previously, this section instantiates them as differentiable spatio-temporal operators to be integrated into the adversarial gradient computation. For an independent local signal segment $\mathbf{x}_{\mathrm{seg}}^{(i)} \in \mathbb{R}^{2 \times L_{\mathrm{seg}}}$ decoupled by the segmentation mechanism (where $L_{\mathrm{seg}}$ is the sample length and the 2 channels represent In-phase and Quadrature components), the algorithm designs specific time-domain and spatial-domain operators.

\subsubsection{Time-Domain Cyclic Shift Operator}
To simulate the synchronization delay characteristics of real hardware systems and block the overfitting of adversarial perturbations to absolute position information, the time-domain cyclic shift operator $T_{\mathrm{time}}(\cdot)$ is applied. The algorithm first sets a maximum shift amount $\tau_{\mathrm{max}}$ based on the synchronization tolerance of typical radar receivers:
\begin{equation}
\tau_{\mathrm{max}} = \max(1, \lfloor \eta \cdot L_{\mathrm{seg}} \rfloor),
\end{equation}
where $\eta$ is the maximum time-shift ratio limiting factor (set to $\eta = 0.02$ by default). Subsequently, an integer shift step $\tau \sim \mathcal{U}(-\tau_{\mathrm{max}}, \tau_{\mathrm{max}})$ is randomly sampled. To strictly maintain the non-mutation of total energy within the observation window, the operator performs a cyclic shift along the time dimension:
\begin{equation}
T_{\mathrm{time}}(\mathbf{x}_{\mathrm{seg}}^{(i)})[c, t] = \mathbf{x}_{\mathrm{seg}}^{(i)}[c, (t - \tau) \bmod L_{\mathrm{seg}}],
\end{equation}
where $c \in \{0, 1\}$ represents the spatial channel index and $t \in [0, L_{\mathrm{seg}}-1]$ represents the time step index.

\subsubsection{Spatial-Domain Amplitude-Phase Distortion Operators}
To simulate the dynamic distortions in signal amplitude and phase induced by complex electromagnetic spaces, we construct the Gaussian noise injection operator $T_{\mathrm{jnr}}(\cdot)$ and the quadrature phase rotation operator $T_{\mathrm{phase}}(\cdot)$. 

For amplitude distortions, the algorithm samples a logarithmic-scale perturbation value $\mathrm{JNR} \sim \mathcal{U}(-5, 5)$ to simulate channel power fluctuations of $\pm 5\,\mathrm{dB}$. Given the average signal power $P_{\mathrm{sig}}$ of the local segment, the required physical noise power is $P_{\mathrm{noise}} = P_{\mathrm{sig}} / 10^{(\mathrm{JNR}/10)}$. A zero-mean Gaussian white noise matrix $\mathbf{w} \sim \mathcal{N}(\mathbf{0}, \sigma_{\mathrm{n}}^2 \mathbf{I})$ is generated, where $\sigma_{\mathrm{n}} = \sqrt{P_{\mathrm{noise}}/2}$. The noise injection operator is expressed as:
\begin{equation}
T_{\mathrm{jnr}}(\mathbf{x}_{\mathrm{seg}}^{(i)}) = \mathrm{Clip}(\mathbf{x}_{\mathrm{seg}}^{(i)} + \mathbf{w}, -1, 1),
\end{equation}
where the $\mathrm{Clip}(\cdot)$ function ensures the amplitude does not exceed the hardware dynamic quantization range of $[-1, 1]$.

For phase distortions, the algorithm randomly samples a rotation angle $\theta \sim \mathcal{U}[0, 2\pi)$ and performs a joint linear transformation on the I/Q channels using a rotation matrix:
\begin{equation}
T_{\mathrm{phase}}(\mathbf{x}_{\mathrm{seg}}^{(i)}) = \begin{bmatrix} \cos\theta & -\sin\theta \\ \sin\theta & \cos\theta \end{bmatrix} \mathbf{x}_{\mathrm{seg}}^{(i)}.
\end{equation}

\subsection{Global-Local Decoupling Segmentation Mechanism}

To interrupt the unconstrained propagation of adversarial gradients across the entire time series and prevent the surrogate network from overfitting to specific global sequence structures (such as absolute positional encodings), SSTA introduces a global-local decoupling segmentation mechanism.

For a given 1D radar baseband signal $\mathbf{x} \in \mathbb{R}^{2 \times L}$, the algorithm randomly samples $N-1$ split points $p_1, p_2, \dots, p_{N-1}$ from the discrete interval $[1, L-1]$. Combined with $p_0 = 0$ and $p_N = L$, the original signal is physically partitioned into $N$ non-overlapping continuous sub-segments:
\begin{equation}
\mathbf{x} = \mathrm{Concat}(\mathbf{x}_{\mathrm{seg}}^{(1)}, \mathbf{x}_{\mathrm{seg}}^{(2)}, \dots, \mathbf{x}_{\mathrm{seg}}^{(N)}),
\end{equation}
where $\mathbf{x}_{\mathrm{seg}}^{(i)} = \mathbf{x}[p_{i-1}:p_i]$, and $\mathrm{Concat}(\cdot)$ is the concatenation operator along the time dimension.

Simultaneously applying extreme amplitude noise and phase rotation to a single local segment can easily lead to the complete loss of its original time-frequency semantics, violating the tactical equivalence constraints. Therefore, SSTA adopts an equiprobable mutually exclusive strategy for spatial-domain distortion processing. The single spatio-temporal joint perturbation process $T_{\mathrm{ST}}(\cdot)$ for each segment is strictly defined as a piecewise function:
\begin{equation}
T_{\mathrm{ST}}(\mathbf{x}_{\mathrm{seg}}^{(i)}) = 
\begin{cases} 
T_{\mathrm{jnr}}\left(T_{\mathrm{time}}(\mathbf{x}_{\mathrm{seg}}^{(i)})\right), & \text{with probability } 0.5, \\ 
T_{\mathrm{phase}}\left(T_{\mathrm{time}}(\mathbf{x}_{\mathrm{seg}}^{(i)})\right), & \text{with probability } 0.5.
\end{cases}
\end{equation}

Finally, the algorithm concatenates all distorted segments strictly in their original temporal order to reconstruct the complete adversarial carrier:
\begin{equation}
T_{\mathrm{seg}}(\mathbf{x}) = \mathrm{Concat}\left(T_{\mathrm{ST}}(\mathbf{x}_{\mathrm{seg}}^{(1)}), T_{\mathrm{ST}}(\mathbf{x}_{\mathrm{seg}}^{(2)}), \dots, T_{\mathrm{ST}}(\mathbf{x}_{\mathrm{seg}}^{(N)})\right).
\end{equation}

This time-slice-based segmented perturbation mechanism forces the surrogate network to extract adversarial features within fine-grained local high-dimensional manifolds, guaranteeing SSTA's outstanding cross-model black-box transferability.

\subsection{Tactical-Equivalent EoT and Momentum Optimization}

To solve the global expected optimization objective formulated in Section \ref{sec2}, the algorithm employs the Monte Carlo method for approximation due to the intractability of analytical solutions in the continuous physical space $\mathcal{T}$. In the $k$-th iteration, for the current adversarial example $\mathbf{x}_{\mathrm{adv}}^{(k)}$, the algorithm executes $T$ independent segmented spatio-temporal transformations in parallel. The aggregated expected adversarial gradient $\mathbf{g}^{(k)}$ is calculated as follows:
\begin{equation}
\mathbf{g}^{(k)} = \frac{1}{T} \sum_{m=1}^{T} \nabla_{\mathbf{x}_{\mathrm{adv}}^{(k)}} J\left(f_{\mathrm{src}}\left(T_{\mathrm{seg},m}(\mathbf{x}_{\mathrm{adv}}^{(k)})\right), y\right).
\end{equation}

By sampling expected gradients within extreme physical boundaries, the generated adversarial perturbations are finely constrained to the low-sensitivity regions of traditional baseband processing, guaranteeing physical transparency to radar matched filtering.

Subsequently, to stabilize the update direction and escape local extrema, the algorithm utilizes the Momentum Iterative Method. Incorporating $L_1$ norm gradient normalization, the momentum $\mathbf{m}^{(k+1)}$ update formula is:
\begin{equation}
\mathbf{m}^{(k+1)} = \mu \cdot \mathbf{m}^{(k)} + \frac{\mathbf{g}^{(k)}}{\|\mathbf{g}^{(k)}\|_1},
\end{equation}
where $\mu$ is the momentum decay factor. Finally, the update direction is extracted using the sign function ($\mathrm{sign}$) and superimposed onto the current adversarial example with step size $\alpha$, followed by $L_\infty$ norm clipping to ensure stealthiness and dynamic range compliance:
\begin{equation}
\mathbf{x}_{\mathrm{adv}}^{(k+1)} = \mathrm{Clip}_{\mathbf{x}, \epsilon} \left( \mathbf{x}_{\mathrm{adv}}^{(k)} + \alpha \cdot \mathrm{sign}(\mathbf{m}^{(k+1)}) \right).
\end{equation}

Through $K$ momentum iteration loops, SSTA outputs adversarial examples that possess both high black-box transferability and strict radar tactical equivalence.

Synthesizing the global-local feature decoupling segmentation mechanism, physical prior-driven operators, and tactical-equivalent EoT strategy, the complete execution procedure of SSTA is summarized in Algorithm \ref{alg:ssta}.

\begin{algorithm}[htbp]
\caption{Momentum Iterative Radar Adversarial Attack Based on SSTA}
\label{alg:ssta}
\SetAlgoNoLine 
\DontPrintSemicolon

\KwIn{Source model $f_{\mathrm{src}}$, loss function $J$;
Clean 1D baseband signal $\mathbf{x} \in \mathbb{R}^{2 \times L}$, true label $y$; Max perturbation bound $\epsilon$; Step size $\alpha$; Iterations $K$; Momentum decay $\mu$; EoT samplings $T$; Segments $N$.}
\KwOut{Adversarial example $\mathbf{x}_{\mathrm{adv}}$.}

\textbf{Initialize:} $\mathbf{x}_{\mathrm{adv}}^{(0)} = \mathbf{x}$, $\mathbf{m}^{(0)} = \mathbf{0}$\;
\For{$k = 0$ \KwTo $K - 1$}{
    $\mathbf{g}_{\mathrm{acc}} = \mathbf{0}$\;
    \For{$m = 1$ \KwTo $T$}{
        Randomly generate $N-1$ split points to divide $\mathbf{x}_{\mathrm{adv}}^{(k)}$ into $N$ local segments $\{\mathbf{x}_{\mathrm{seg},m}^{(1)}, \dots, \mathbf{x}_{\mathrm{seg},m}^{(N)}\}$\;
        \For{$j = 1$ \KwTo $N$}{
            $\tilde{\mathbf{x}}_{\mathrm{seg},m}^{(j)} = T_{\mathrm{time}}(\mathbf{x}_{\mathrm{seg},m}^{(j)})$\;
            $r \sim \mathcal{U}(0, 1)$\;
            \eIf{$r < 0.5$}{
                $\hat{\mathbf{x}}_{\mathrm{seg},m}^{(j)} = T_{\mathrm{jnr}}(\tilde{\mathbf{x}}_{\mathrm{seg},m}^{(j)})$\;
            }{
                $\hat{\mathbf{x}}_{\mathrm{seg},m}^{(j)} = T_{\mathrm{phase}}(\tilde{\mathbf{x}}_{\mathrm{seg},m}^{(j)})$\;
            }
        }
        $T_{\mathrm{seg}, m}(\mathbf{x}_{\mathrm{adv}}^{(k)}) = \mathrm{Concat}(\hat{\mathbf{x}}_{\mathrm{seg},m}^{(1)}, \dots, \hat{\mathbf{x}}_{\mathrm{seg},m}^{(N)})$\;
        $\mathbf{g}_m = \nabla_{\mathbf{x}_{\mathrm{adv}}} J\left(f_{\mathrm{src}}\left(T_{\mathrm{seg}, m}(\mathbf{x}_{\mathrm{adv}}^{(k)})\right), y\right)$\;
        $\mathbf{g}_{\mathrm{acc}} = \mathbf{g}_{\mathrm{acc}} + \mathbf{g}_m$\;
    }
    $\mathbf{g}^{(k)} = \mathbf{g}_{\mathrm{acc}} / T$\;
    $\mathbf{m}^{(k+1)} = \mu \cdot \mathbf{m}^{(k)} + \mathbf{g}^{(k)} / \|\mathbf{g}^{(k)}\|_1$\;
    $\tilde{\mathbf{x}}_{\mathrm{adv}} = \mathbf{x}_{\mathrm{adv}}^{(k)} + \alpha \cdot \mathrm{sign}(\mathbf{m}^{(k+1)})$\;
    $\mathbf{x}_{\mathrm{adv}}^{(k+1)} = \mathrm{Clip}_{\mathbf{x}, \epsilon}(\tilde{\mathbf{x}}_{\mathrm{adv}})$\;
}
\Return{$\mathbf{x}_{\mathrm{adv}}^{(K)}$}\;
\end{algorithm}

\section{Experiments and Results Analysis} \label{sec4}

This section details the experimental infrastructure, comparative baselines, and quantitative metrics. To address the complexity of radar electronic warfare scenarios and prevent dimension explosion and attribution confusion in subsequent cross-evaluations, a rigorous representative sampling and evaluation logic is established at the outset of the experimental design.

\subsection{Experimental Setup and Evaluation Metrics}

This section details the experimental infrastructure, comparative baselines, and quantitative metrics. To address the complexity of radar electronic warfare scenarios and prevent dimension explosion and attribution confusion in subsequent cross-evaluations, a rigorous representative sampling and evaluation logic is established at the outset of the experimental design.

\subsubsection{Dataset Construction}
Because real military radar engagement data is highly sensitive and classified, this experiment constructs a radar jamming signal simulation dataset based on typical active jamming systems. The radar transmitted waveform employs a Linear Frequency Modulation (LFM) signal with a pulse width of $20~\mu\mathrm{s}$ and a bandwidth of $10~\mathrm{MHz}$. The dataset comprises 10 typical jamming types (covering 6 types of single jamming and 4 types of composite jamming, detailed in Table \ref{tab:jamming_types}). The simulation dataset consists of 1,000 independent randomly generated samples for each type. All signals extract the complex baseband envelopes of the $\mathrm{I}/\mathrm{Q}$ dual channels, with the sampling length truncated to 2000, and are Min-Max normalized to $[-1, 1]$ to accommodate the physical dynamic range of the digital baseband.


\begin{table}[htbp]
  \centering
  \renewcommand\arraystretch{1.5}
  \caption{Summary of Jamming Signals in the Simulation Dataset}
    \resizebox{\linewidth}{!}{
    \begin{tabular}{p{4.875em}p{11.125em}p{10.835em}}
    \toprule
    Jamming Type & Key Parameter Range & Modulation\newline{}Characteristics \\
    \midrule
    DFTJ  & False targets: 3--6, \newline{}Delay: 1--10~$\mu\text{s}$ & Multi-pulse linear\newline{}time shifting \\
    ISRJ  & Retransmissions: 1--4, \newline{}Sampling period: 5--20~$\mu\text{s}$ & Periodic spectral\newline{} notches \\
    RDJ   & Lag time: 1--10~$\mu\text{s}$ & Constant frequency shift\newline{}+ time delay extension \\
    SNJ   & Duty cycle: 20\%--50\%, \newline{}Delay: 1--5~$\mu\text{s}$ & Intermittent sampled\newline{}noise retransmission \\
    SPJ   & Bandwidth: 20--40~MHz & Narrowband spectrum\newline{}dwelling \\
    SWJ   & Sweep period: 40--80~$\mu\text{s}$,\newline{}Bandwidth: 20~MHz & Time-varying sloped\newline{}sawtooth spectrum \\
    Composite Jamming & Superposition of 2--3 \newline{}jamming parameters & Multi-modal time-\newline{}frequency coupling \\
    \bottomrule
    \end{tabular}}%
  \label{tab:jamming_types}%
\end{table}%

\subsubsection{Evaluation Model Pool and Representative Sampling}
To investigate the generalization laws of adversarial perturbations across different underlying feature extraction mechanisms, this paper selects 9 highly representative heterogeneous networks as evaluation baselines. The model pool comprehensively covers three mainstream architectures: CNNs based on local receptive fields (ResNet18 \cite{He2016ResNet}, DenseNet121 \cite{Huang2017DenseNet}, EfficientNet \cite{Tan2019EfficientNet}, SEResNet \cite{Hu2018SENet}), RNNs focusing on long-range hidden state accumulation (DeepLSTM \cite{Hochreiter1997LSTM}, DeepGRU \cite{Cho2014GRU}, CNN-BiLSTM \cite{Sainath2015CNNBLSTM}), and Transformers based on global self-attention (SwinTransformer \cite{Liu2021Swin}, Conformer \cite{Gulati2020Conformer}). To manage the computational overhead of the Expectation over Transformation (EoT), we uniformly sample 100 samples from each jamming category to construct a standardized adversarial evaluation subset totaling 1,000 samples. On this subset, the defense systems exhibit extremely high classification accuracies (exceeding 98.00\% across all models, with six of them achieving exactly 100\%), providing a rigorous baseline for adversarial testing.

Faced with the potential spatial evaluation dimension explosion of a $9 \times 9$ transfer matrix, this paper adopts a rigorous representative sampling logic for subsequent mechanism and physical robustness analyses. Specifically, ResNet18 is identified as the Anchor Surrogate. As a benchmark in deep learning, its local receptive field characteristics provide a rigorous reference base for evaluating cross-architecture transfer. To maximize architectural differences and evaluation stringency, specific black-box targets are selected based on the principle of Orthogonal Inductive Biases: SwinTransformer and Conformer represent global attention paradigms, DeepGRU represents gated sequential memory mechanisms, whereas DenseNet and EfficientNet represent advanced dense topologies. This orthogonal combination can effectively examine the underlying generalization capability of adversarial perturbations. Furthermore, to avoid visual clutter in mechanism visualization and physical validation, a core evaluation subset is selected from the eight comparative baselines: the traditional momentum baseline MI-FGSM, the input transformation SIA, the strong baseline VMI-FGSM, and the proposed SSTA.

\subsubsection{Evaluation Metrics and Implementation Details}
The evaluation framework of this experiment is built upon the axiomatic logic of ``fundamental attack success rate---adversarial space division---statistical derivation.'' Let the standardized adversarial evaluation subset be $\mathcal{D}_{\mathrm{test}} = \{(\mathbf{x}_i, y_i)\}_{i=1}^{N}$, and the adversarial samples generated by the source model $f_{\mathrm{src}}$ be $\{\mathbf{x}_{\mathrm{adv}, i}^{\mathrm{src}}\}_{i=1}^{N}$. For a given target model $f_{\mathrm{tgt}}$, the foundational Attack Success Rate (ASR) is defined as the proportion of samples causing misclassification:
\begin{equation} \label{eq:asr}
\mathrm{ASR}(f_{\mathrm{src}} \rightarrow f_{\mathrm{tgt}}) = \frac{1}{N} \sum_{i=1}^{N} \mathbb{I} \left( \arg\max f_{\mathrm{tgt}}(\mathbf{x}_{\mathrm{adv}, i}^{\mathrm{src}}) \neq y_i \right)
\end{equation}
where $\mathbb{I}(\cdot)$ is the indicator function. Upon this cornerstone, according to the attack-defense boundary, the evaluation system is strictly divided into two core adversarial spaces:
\begin{enumerate}
    \item \textbf{White-Box ASR ($\mathrm{WB\text{-}ASR}$):} Constrained by $f_{\mathrm{src}} = f_{\mathrm{tgt}}$, reflecting the absolute attack strength against the source model's features.
    \item \textbf{Black-Box ASR ($\mathrm{BB\text{-}ASR}$):} Constrained by $f_{\mathrm{src}} \neq f_{\mathrm{tgt}}$, representing the underlying transfer capability across heterogeneous network feature barriers.
\end{enumerate}

Subsequent performance metrics are derived from these statistics, including Average WB-ASR (AWB-ASR), Average BB-ASR (ABB-ASR), Cross-Architecture BB-ASR, and Generalization Gap. Furthermore, addressing the distortion of radar signals in physical channels, an Effective ASR metric is specifically introduced in Section 4.5.

All gradient-based iterative attack algorithms follow identical underlying constraints: the infinity norm is strictly limited to $\epsilon = 0.05$, the single iteration step size is $\alpha = 0.005$, iterations $K = 20$, and momentum decay factor $\mu = 1.0$. For the proposed SSTA, unique hyperparameters are optimally configured based on sensitivity analysis: segment count $N = 3$, EoT physical sampling count $T = 10$, and max delay ratio limit $\eta = 0.02$.

\subsection{Comprehensive Attack Performance and Transferability}
This section presents and analyzes the black-box transfer attack results of the proposed algorithm and the comparison baselines on 9 heterogeneous target models. By quantitatively analyzing the cross-transfer success rates of adversarial samples between heterogeneous cognitive systems, from two dimensions of the macroscopic black-box space's overall performance and the microscopic cross-architecture transfer links, we explore the generalization rules of adversarial perturbations across the underlying feature extraction mechanisms.

\subsubsection{Comprehensive Black-Box Space Performance}

In the macroscopic evaluation dimension, based on the standardized adversarial evaluation subset of 1,000 samples established in Section IV-A, this experiment comprehensively records the transfer performance of adversarial samples generated by various attack algorithms on the full set of models against unknown black-box target models. Table \ref{tab:comprehensive_tasr} details the global black-box transfer attack success rates.

\begin{table*}[htbp]
\centering
\caption{Comprehensive Black-Box Transfer Attack Success Rates (\%) Across 9 Heterogeneous Models}
\label{tab:comprehensive_tasr}
\renewcommand{\arraystretch}{1.3}
\setlength{\tabcolsep}{4.5pt}
\resizebox{\textwidth}{!}{%
\begin{tabular}{@{} l ccccccccc c @{}}
\toprule
Algorithm & ResNet18 & DenseNet121 & EfficientNet & SEResNet & DeepLSTM & DeepGRU & CNN-BiLSTM & SwinTrans & Conformer & Average BB-ASR \\
\midrule
FGSM \cite{Goodfellow2015Explaining} & 34.89 & 18.29 & 45.71 & 41.34 & 51.46 & 46.20 & 36.06 &  \underline{38.62} & 51.81 & 40.49 \\
MI-FGSM \cite{Dong2018Boosting} & 47.79 & 24.50 & 39.04 & 55.62 & 50.19 & 49.86 & 32.58 & 33.24 & 59.64 & 43.61 \\
VMI-FGSM \cite{Xiaosen2021} &  \underline{63.80} &  \underline{62.44} & 44.69 &  \underline{63.98} &  \textbf{61.01} &  \textbf{66.39} &  \underline{43.56} & 32.60 &  \underline{71.40} &  \underline{56.65} \\
DIM \cite{Xie2018Improving} & 49.54 & 22.81 & 50.02 & 55.95 & 43.33 & 53.81 & 24.98 & 33.62 & 65.25 & 44.37 \\
TIM \cite{Yinpeng2019Evading} & 28.59 & 13.78 & 31.52 & 31.48 & 30.39 & 28.48 & 27.57 & 38.59 & 42.58 & 30.33 \\
SIA \cite{Wang2023Structure} & 53.69 & 48.84 &  \underline{51.78} & 57.92 & 53.01 & 62.60 & 33.10 & 35.60 & 52.55 & 49.90 \\
ADMIX \cite{WangKun2021Admix} & 50.96 & 34.49 & 40.16 & 28.76 & 34.16 & 46.73 & 31.46 & 33.53 & 40.58 & 37.87 \\
FTM \cite{Wang2024Improving} & 49.56 & 6.61 & 45.04 & 50.02 & 28.03 & 39.05 & 20.62 & 37.01 & 58.21 & 37.13 \\
\textbf{SSTA (Ours)} & \textbf{76.92} & \textbf{73.31} & \textbf{66.55} & \textbf{80.01} & \underline{60.26} & \underline{65.81} & \textbf{50.25} & \textbf{56.74} & \textbf{73.19} & \textbf{67.01} \\
\bottomrule
\end{tabular}%
}
\end{table*}

Statistical data indicate that the SSTA algorithm exhibits a significant advantage in comprehensive transfer capability. Its Average BB-ASR reaches 67.01\%, achieving an absolute improvement of 10.36\% compared to the best-performing baseline algorithm VMI-FGSM (56.65\%). Our analysis reveals that traditional adversarial attack methods based on the visual priors of 2D images exhibit notable limitations when applied to 1D radar baseband signals. For instance, TIM and FTM yield Average BB-ASRs of only 30.33\% and 37.13\%, respectively. This is because the local temporal features and global phases of radar sequences possess strong physical coupling. Simple 1D spatial smoothing or feature tuning easily destroys the complex envelope coherent structure of the perturbed waveform, leading to severe source-model overfitting. In contrast, SSTA mitigates this by introducing spatio-temporal dual-domain physical operators aligned with receiver mechanisms, effectively enhancing the robustness of the perturbation under black-box transfer conditions.

\begin{figure*}[htbp]
    \centering
    \includegraphics[width=\textwidth]{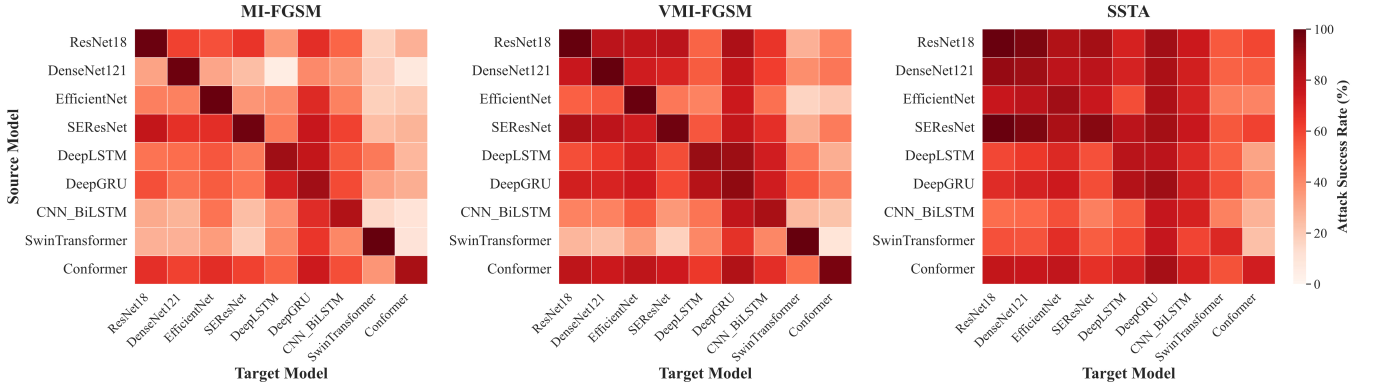}
    \caption{Heatmap of transfer attack success rates across the $9 \times 9$ cross-evaluation matrix.}
    \label{fig:heatmap}
\end{figure*}

To more intuitively visualize the distribution patterns of the global transfer performance, Fig. \ref{fig:heatmap} presents the attack success rate heatmaps for MI-FGSM, VMI-FGSM, and SSTA under the $9 \times 9$ cross-evaluation matrix. In the color mapping of this experiment, darker tones represent higher attack success rates, while lighter tones indicate lower success rates. Visually, the dark areas of MI-FGSM are almost entirely concentrated along the diagonal, indicating its weak black-box generalization capability; VMI-FGSM shows slightly deepened tones on some near-source models, but its overall performance across structurally distinct networks remains pale. Conversely, SSTA exhibits broad and uniform dark distributions in off-diagonal regions, confirming that the proposed algorithm effectively alleviates the reliance of adversarial perturbations on specific source network parameters. Notably, even when utilizing SwinTransformer (which relies heavily on global absolute positional encoding) as the source model---where traditional algorithms struggle to transfer because the gradients contain extensive specialized information targeting global correlations---SSTA still maintains a transfer success rate lower bound of 56.74\%. This value even outperforms the migration performance of most traditional algorithms under optimal source model conditions, verifying the generalization stability of the algorithm independent of specific high-quality surrogate models.

\subsubsection{Cross-Architecture Feature Barrier Breakthrough Analysis}

\begin{figure}[htbp]
    \centering
    \includegraphics[width=0.8\columnwidth]{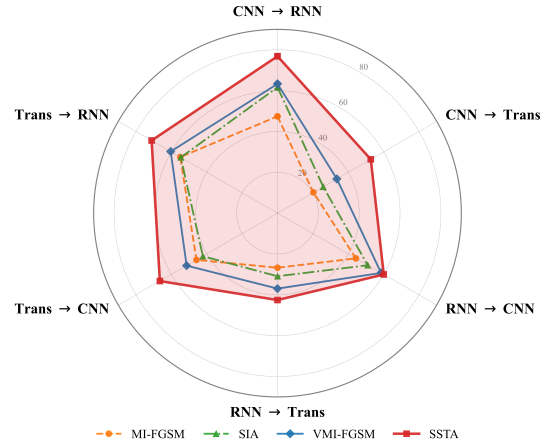}
    \caption{Radar chart of cross-architecture transfer attack success rates on six core paths.}
    \label{fig:radar}
\end{figure}

Maximizing the architectural orthogonality (Orthogonal Inductive Biases) of cognitive defense systems is a core defensive measure to block adversarial generalization. To deeply investigate the breakthrough mechanism of SSTA across different feature aggregation paradigms, conditional averaging is performed on model pairs belonging to the same architecture based on the full evaluation matrix (Table \ref{tab:comprehensive_tasr}). Based on this calculation logic, Table \ref{tab:cross_arch_tasr} and Fig. \ref{fig:radar} present the fine-grained quantitative results and the cross-architecture transfer radar chart of the core comparative subset (including MI-FGSM, SIA, VMI-FGSM, and SSTA) on 6 cross-architecture backbone transfer links.

\begin{table}[htbp]
\centering
\caption{Transfer Attack Success Rates (\%) Across Different Feature Extraction Mechanisms}
\label{tab:cross_arch_tasr}
\renewcommand{\arraystretch}{1.3} 
\setlength{\tabcolsep}{4pt}       
\begin{tabular}{@{} l cccc @{}}
\toprule
Transfer Path & MI-FGSM & SIA & VMI-FGSM & SSTA \\
\midrule
CNN$\rightarrow$RNN   & 47.41 & 61.66 & 63.34 & \textbf{76.85} \\
CNN$\rightarrow$Trans & 20.30 & 25.85 & 33.55 & \textbf{52.73} \\
RNN$\rightarrow$CNN   & 44.31 & 51.02 & 58.55 & \textbf{60.07} \\
RNN$\rightarrow$Trans & 26.72 & 30.90 & 36.95 & \textbf{42.55} \\
Trans$\rightarrow$CNN & 45.81 & 42.18 & 51.38 & \textbf{66.50} \\
Trans$\rightarrow$RNN & 54.95 & 54.65 & 60.37 & \textbf{71.17} \\
\bottomrule
\end{tabular}
\end{table}

The geometric profile of the radar chart visually reflects the asymmetric blocking characteristics of deep learning architectural differences against adversarial energy. On paths where feature extraction preferences partially intersect (e.g., the transition from local convolutions to temporal memory accumulation in the $\mathrm{CNN} \rightarrow \mathrm{RNN}$ path, and from global attention to temporal memory in the $\mathrm{Trans} \rightarrow \mathrm{RNN}$ path), the core algorithms generally exhibit favorable transferability. Specifically, SSTA's black-box transfer success rates reach $76.85\%$ and $71.17\%$, respectively, indicating that introducing physical channel priors via Expectation over Transformation significantly broadens the feature coverage of adversarial examples within the high-dimensional manifold space.

However, when the transfer link shifts to the chasm crossing from local receptive field architectures to global self-attention architectures (i.e., the $\mathrm{CNN} \rightarrow \mathrm{Trans}$ and $\mathrm{RNN} \rightarrow \mathrm{Trans}$ paths), traditional algorithms exhibit widespread limitations in attack performance due to their inability to overcome the underlying feature aggregation barriers. As illustrated in Fig. \ref{fig:radar}, the contour lines of the baseline algorithms shrink sharply inward when facing the two Transformer target axes. Notably, the transfer success rates of MI-FGSM and SIA on the $\mathrm{CNN} \rightarrow \mathrm{Trans}$ path drop to as low as $20.30\%$ and $25.85\%$, respectively. Even the strong baseline VMI-FGSM, which incorporates variance tuning, can only maintain transfer levels of $33.55\%$ and $36.95\%$ on these two paths.

In stark contrast, SSTA's performance envelope achieves a significant outward expansion on these two arduous paths, remarkably securing transfer success rates of $52.73\%$ ($\mathrm{CNN} \rightarrow \mathrm{Trans}$) and $42.55\%$ ($\mathrm{RNN} \rightarrow \mathrm{Trans}$) against the trend. This result validates the effectiveness of the global-local feature decoupling segmentation mechanism: by introducing random physical slice boundaries along the time axis, the algorithm forcefully interrupts the adversarial gradient propagation's over-reliance on the long-range coherent features of a single network, avoiding the perturbation's local overfitting to the source network's absolute positional encodings, and thereby effectively penetrating the multi-head attention defense mechanisms of the black-box Transformer. Furthermore, the closed red curve of SSTA in the radar chart is relatively balanced across all axes, eliminating the asymmetric fluctuation characteristics of traditional algorithms that heavily depend on specific transfer directions. This fundamentally consolidates the tactical dominance of the perturbed waveforms under unknown orthogonal architecture defenses from the perspective of underlying feature optimization dynamics.

\subsection{Ablation Study and Parameter Sensitivity}

This section aims to verify the necessity of the spatio-temporal dual-domain operators and their cascading strategy in SSTA through ablation experiments, and to deeply explore the underlying effects of key hyperparameters---such as the number of iterations, the number of segments, and the sampling frequency---on the model's convergence and generalization performance.

\subsubsection{Ablation Analysis of Spatio-Temporal Operators}

\begin{figure}[htbp]
    \centering
    \includegraphics[width=\columnwidth]{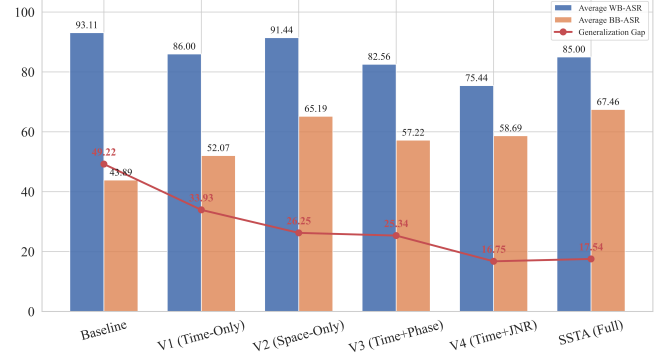}
    \caption{Performance comparison trends of different ablation variants.}
    \label{fig:ablation}
\end{figure}

To quantify the independent contributions of each physical operator to the model's feature decoupling and generalization capabilities, this experiment constructs 4 independent algorithm variants on the standardized test subset, using the traditional MI-FGSM as the baseline algorithm (Baseline) and progressively stripping or replacing the core physical modules in SSTA. Specifically, variant V1 retains only the time-shift operator; V2 uses only the spatial-domain operators (i.e., equiprobably applying JNR noise or phase rotation); V3 employs a cascade of time shift and a single phase rotation operator; and V4 employs a cascade of time shift and a single JNR noise injection operator. Table \ref{tab:ablation} records the precise quantitative evaluation metrics for each variant. Meanwhile, Fig. \ref{fig:ablation} plots an intuitive performance comparison bar-line chart, where the bars represent the average attack success rate and the red line represents the Generalization Gap (GAP).

\begin{table}[htbp]
\centering
\caption{Performance Comparison of Different Ablation Variants}
\label{tab:ablation}
\renewcommand{\arraystretch}{1.3}
\resizebox{\columnwidth}{!}{%
\begin{tabular}{@{} l ccc @{}}
\toprule
Variant & AWB-ASR (\%) & ABB-ASR (\%) & Gap (\%) \\
\midrule
Baseline (MI-FGSM)   & 93.11          & 43.89          & 49.22          \\
V1: Time-Only        & 86.00          & 52.07          & 33.93          \\
V2: Space-Only       & 91.44          & 65.19          & 26.25          \\
V3: Time+Phase       & 82.56          & 57.22          & 25.34          \\
V4: Time+JNR         & 75.44          & 58.69          & 16.75          \\
\textbf{SSTA (Full)} & \textbf{85.00} & \textbf{67.46} & \textbf{17.54} \\
\bottomrule
\end{tabular}%
}
\end{table}

From the experimental results, it can be observed that the purely digital domain baseline algorithm (Baseline), which completely lacks physical priors, falls into severe white-box overfitting, with its Generalization Gap (GAP) reaching as high as $49.22\%$. After introducing single-domain physical operators (V1, V2), the overfitting phenomenon is alleviated to a certain extent, and the GAP narrows significantly. However, unrestrictedly enforcing the superposition of cross-domain operators (V3, V4) contrarily leads to the overload distortion of adversarial energy. For instance, the Average White-Box Success Rate (AWB-ASR) of V4 drops sharply to $75.44\%$, indicating that excessive joint transformations severely destroy the fundamental features of the original deception jamming, resulting in the degradation of the algorithm's foundational blinding lethality. The SSTA algorithm successfully resolves this contradiction by ingeniously introducing an ``equiprobable mutually exclusive'' cascading strategy for spatial operators. While stabilizing the AWB-ASR at $85.00\%$, it pushes the Average BB-ASR to a peak of $67.46\%$. This result not only proves the indispensability of spatio-temporal dual-domain operators in resisting black-box defenses but also confirms the scientific rationality of the SSTA cascading strategy in balancing ``underlying feature destruction'' and ``generalization space expansion''.

\subsubsection{Sensitivity to Iteration Count and Convergence}

\begin{figure}[htbp]
    \centering
    \includegraphics[width=\columnwidth]{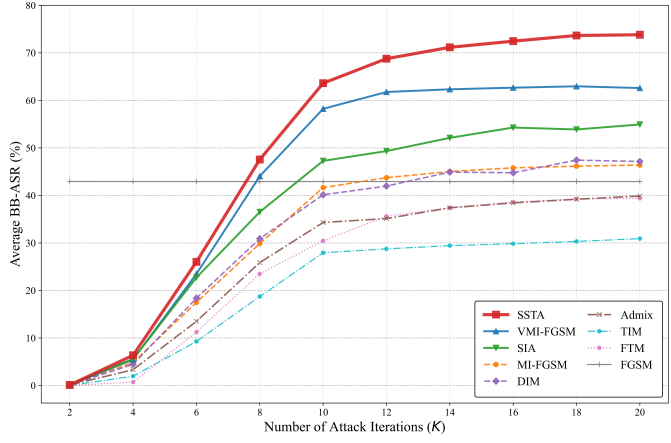}
    \caption{Evolution curves of Average BB-ASR with respect to the iteration number $K$.}
    \label{fig:convergence}
\end{figure}

The number of momentum iterations $K$ determines the optimization depth of adversarial perturbations in the solution space. Fig. \ref{fig:convergence} tracks the evolutionary trajectories of the Average BB-ASR across 72 cross-evaluation paths with respect to the iteration number $K \in [0, 20]$. It can be observed that most traditional momentum-based algorithms tend to saturate within the interval of $K=14$ to $K=16$, subsequently entering a plateau or even exhibiting slight degradation (overfitting). In contrast, SSTA's performance curve consistently overrides all comparative algorithms throughout the entire iteration cycle and maintains a steadily rising performance scaling trend even at $K=20$. This indicates that the expectation over transformation guided by physical priors not only smooths the loss manifold but also effectively avoids gradient oscillation and stagnation at local minima, endowing the algorithm with stronger resilience against overfitting and deeper feature generalization potential.

\subsubsection{Sensitivity to Segmentation Count and Sampling Frequency}

The number of segments $N$ in the global-local feature decoupling mechanism directly determines the blocking granularity of adversarial gradients along the time axis. As can be seen from the macroscopic data in Table \ref{tab:seg_N}, when physical slicing is not performed ($N=1$), globally uniform distortions fail to break the model's dependence on coherent features, resulting in an Average BB-ASR of merely $47.26\%$. As the number of segments increases to $N=3$, the Average BB-ASR experiences a leap. However, when $N \ge 5$, the performance growth exhibits a clear diminishing marginal utility effect. 

\begin{table}[htbp]
\centering
\caption{Attack Success Rates (\%) Under Different Segmentation Counts $N$}
\label{tab:seg_N}
\renewcommand{\arraystretch}{1.3}
\begin{tabularx}{\columnwidth}{@{} l *{6}{>{\centering\arraybackslash}X} @{}}
\toprule
\textbf{Segments ($N$)} & \textbf{1} & \textbf{2} & \textbf{3} & \textbf{5} & \textbf{7} & \textbf{10} \\
\midrule
AWB-ASR & 64.33 & 81.22 & 85.00 & 85.78 & 87.00 & 87.56 \\
ABB-ASR & 47.26 & 64.31 & 67.46 & 68.68 & 69.61 & 69.71 \\
\bottomrule
\end{tabularx}
\end{table}

\begin{figure}[htbp]
    \centering
    \includegraphics[width=\columnwidth]{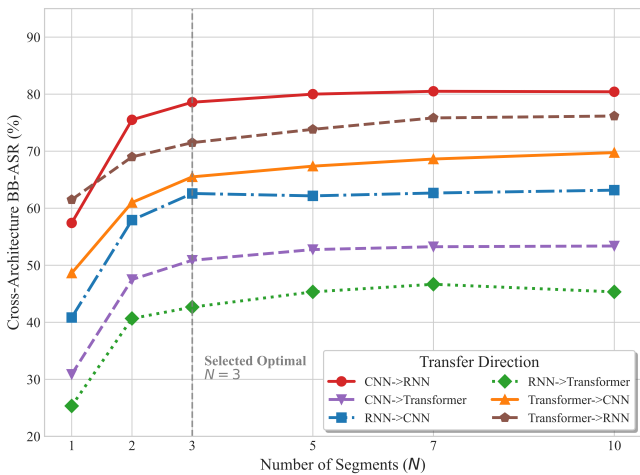}
    \caption{Evolution dynamics of transfer success rates across different architectural paths with respect to segmentation count $N$.}
    \label{fig:seg_N}
\end{figure}

The cross-architecture curves in Fig. \ref{fig:seg_N} profoundly reveal the microscopic decoupling mechanism behind this phenomenon: when $N$ changes from 1 to 3, the curves of all 6 cross-domain links (including the $\mathrm{CNN} \rightarrow \mathrm{Trans}$ and $\mathrm{RNN} \rightarrow \mathrm{Trans}$ paths, which possess the deepest feature chasms) exhibit extremely steep climbs. This demonstrates that local segmentation forcefully interrupts the overfitting of the Transformer's multi-head attention mechanism to long-range absolute positional encodings. Once crossing the $N=3$ mark, the slopes of all curves rapidly flatten, indicating that the gradient decoupling advantage has been fully released. Considering that excessive slicing leads to signal fragmentation and destroys the physical coherence of actual radar pulse modulation, this experiment comprehensively balances generalization gains and waveform fidelity, strictly establishing the optimal segmentation count as $N=3$.

\begin{table}[htbp]
\centering
\caption{Performance and Overhead Under Different EoT Sampling Counts $T$}
\label{tab:eot_T}
\renewcommand{\arraystretch}{1.3}
\resizebox{\columnwidth}{!}{%
\begin{tabular}{@{} l ccccc @{}}
\toprule
\textbf{EoT Samplings ($T$)} & \textbf{1} & \textbf{5} & \textbf{10} & \textbf{15} & \textbf{20} \\
\midrule
AWB-ASR (\%) & 77.67 & 82.00 & 85.00 & 84.33 & 85.44 \\
ABB-ASR (\%) & 59.86 & 65.63 & 67.46 & 68.63 & 69.39 \\
Total Time (s) & 495.75 & 544.93 & 720.61 & 1158.04 & 1550.88 \\
Avg Time (s) & 0.55 & 0.61 & 0.80 & 1.29 & 1.72 \\
\bottomrule
\end{tabular}%
}
\end{table}

\begin{figure*}[htbp]
    \centering
    \includegraphics[width=\textwidth]{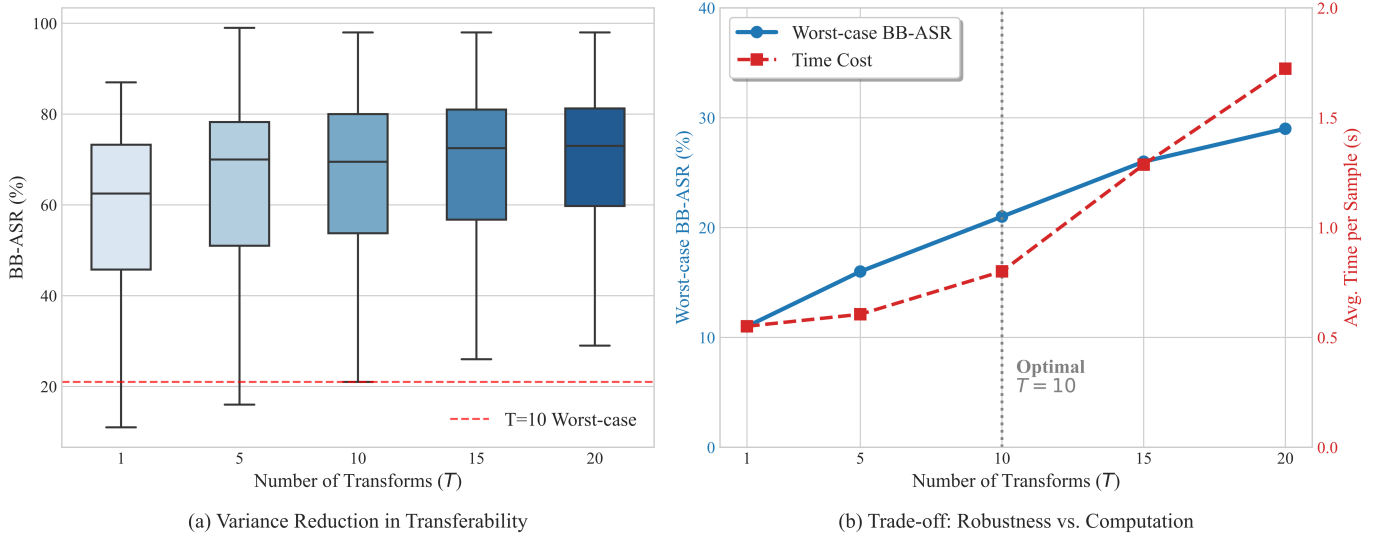}
    \caption{Comprehensive evaluation of EoT sampling count $T$: (a) Boxplot of cross-model transfer rate variance; (b) Trade-off between worst-case transfer rate and time cost.}
    \label{fig:eot_T}
\end{figure*}

On the other hand, the Expectation over Transformation sampling count $T$ is the key to balancing generalization stability and computational overhead. Based on the aforementioned standardized test sample subset, Table \ref{tab:eot_T} records the Average BB-ASR and the average generation time per sample as the number of samplings varies. Fig. \ref{fig:eot_T} provides a set of in-depth dual evaluation subplots: the left subplot \ref{fig:eot_T}(a) is a boxplot of the black-box transfer success rates (displaying variance and extreme cases), while the right subplot \ref{fig:eot_T}(b) shows the trade-off curve between the generalization lower bound and time cost. Combined with the distribution evolution in Fig. \ref{fig:eot_T}(a), a single blind sampling ($T=1$) leads to extremely unstable gradient update directions. This is visually reflected by the exceptionally wide interquartile range in the boxplot, with the worst-case transfer rate dropping close to 10\%. As $T$ increases to 10, the boxes converge significantly, effectively achieving variance reduction and solidly elevating the worst-case lower bound to above 20\%. However, Fig. \ref{fig:eot_T}(b) reveals the computational bottleneck of unrestrictedly increasing $T$. Beyond $T=10$, the blue solid line representing the generalization lower bound exhibits a logarithmic deceleration trend (yielding marginal gains up to $T=20$), whereas the red dashed line representing the time cost surges almost linearly. As detailed in Table \ref{tab:eot_T}, forcibly increasing $T$ from 10 to 20 doubles the average time cost per sample from $0.80\ \mathrm{s}$ to $1.72\ \mathrm{s}$. In radar electronic warfare scenarios with stringent latency constraints, such a duplication in time cost is unacceptable. Therefore, synthesizing the intersecting trends of the curve slopes in Fig. \ref{fig:eot_T}(b) and practical engineering constraints, this paper identifies $T=10$ as the optimal balance point between high tactical effectiveness and low computational overhead.

\subsection{Mechanistic Analysis of Transferability}

The previous macroscopic evaluations have fully quantified the significant advantages of the SSTA algorithm when crossing heterogeneous cognitive defense systems. To explore the origin of this high transferability from the underlying mathematical principles of deep learning, this section will follow the representative sampling logic established in Section IV-A. By selecting specific source and target model paths, we deeply dissect the microscopic optimization dynamics mechanism of adversarial perturbations breaking through architectural barriers from two core dimensions: high-dimensional Feature Manifold and cross-model Gradient Alignment.

\subsubsection{Evolution of High-Dimensional Feature Manifolds}

\begin{figure*}[htbp]
    \centering
    \includegraphics[width=0.98\textwidth]{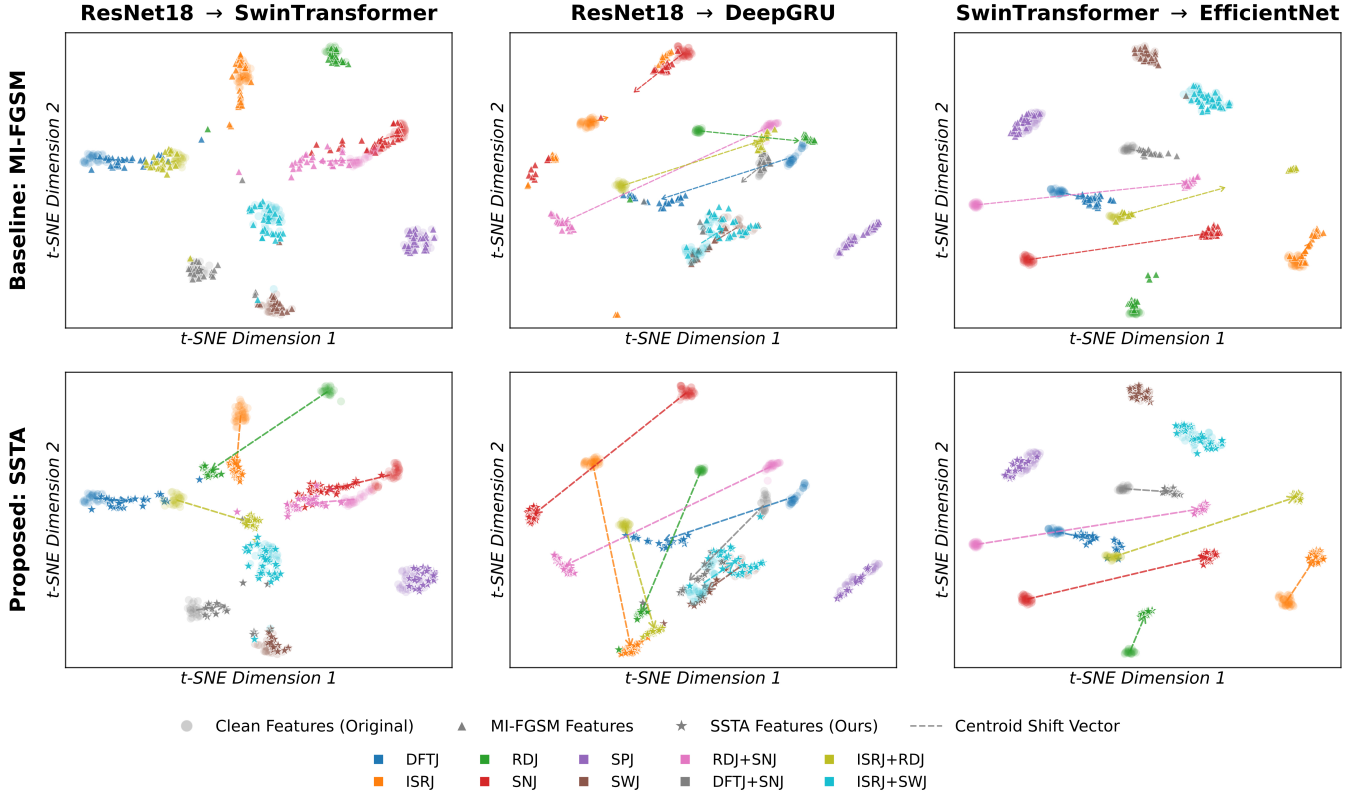}
    \caption{t-SNE dimensionality reduction visualization of hidden feature manifolds under different cross-architecture transfer paths.}
    \label{fig:tsne}
\end{figure*}

The classification decision of deep neural networks essentially partitions hyperplanes in a high-dimensional latent space. For an adversarial example to be effective in a black-box model, the deep feature distribution it extracts within the target model must successfully cross the decision boundary of the original clean samples. To balance the clarity of high-dimensional spatial visualization and the statistical representativeness of categories, this experiment uniformly extracts 300 standardized samples containing all 10 jamming systems from the global test set as the analysis base by fixing the random seed. Based on this, t-SNE technology is utilized to perform dimensionality reduction visualization on the hidden features output by the penultimate layer of the target model. Fig. \ref{fig:tsne} extracts the feature manifold distributions of the baseline algorithm (MI-FGSM) and the proposed algorithm (SSTA) on three cross-architecture physical links ($\mathrm{ResNet18} \rightarrow \mathrm{SwinTransformer}$, $\mathrm{ResNet18} \rightarrow \mathrm{DeepGRU}$, and $\mathrm{SwinTransformer} \rightarrow \mathrm{EfficientNet}$). The scatter points represent clean samples (light-colored dots) and adversarial samples (dark-colored markers), respectively, while the dashed arrows precisely quantify the Centroid Shift Vector for each category.

Observing the high-dimensional manifold distributions in Fig. \ref{fig:tsne}, because MI-FGSM relies heavily on the specific local receptive field of the source model, the adversarial sample features it generates stagnate at the margins of the clean sample clusters in the black-box latent space of the target model. The centroid shift vectors are extremely short, failing to substantively penetrate the discriminant boundary of the defense model. In contrast, SSTA generally induces significant directed centroid shifts, and after breaking away from the original distribution, it re-aggregates into highly compact, erroneous feature clusters.

A deeper analysis of the microscopic evolution trajectories of specific cross-domain links can further reveal the non-uniformity differences in adversarial generalization at the feature level. First, in the $\mathrm{ResNet18} \rightarrow \mathrm{SwinTransformer}$ group with the most drastic architectural differences, the adversarial feature clusters of MI-FGSM almost completely overlap with the clean samples. Conversely, SSTA successfully induces the multi-class jamming features (e.g., RDJ, ISRJ, SNJ, and ISRJ+RDJ) to cross the stringent boundary of the global self-attention mechanism, achieving clear directed shifts. Furthermore, in the $\mathrm{ResNet18} \rightarrow \mathrm{DeepGRU}$ and $\mathrm{SwinTransformer} \rightarrow \mathrm{EfficientNet}$ links, a critical phenomenon of adversarial manifold degradation is revealed: although MI-FGSM compels some samples to escape the original cluster, a large number of adversarial samples remain highly entangled with the clean distribution, leading to a severe ``intra-class dispersion'' of the adversarial cluster in the latent space. This drastic intra-class divergence causes its arithmetic centroid shift vector to actually point to an empty manifold region lacking genuine sample support (i.e., an ineffective discriminant domain). In stark contrast, the adversarial samples generated by SSTA not only successfully detach from the original distribution but also maintain extremely high feature cohesion in the unknown black-box space, thereby forming a far more lethal and directed misleading clustering.

However, from a rigorous academic perspective, the above experiments also objectively expose the inherent physical limitations of the current additive adversarial mechanism. Observing Fig. \ref{fig:tsne}, it can be found that for categories such as sweep jamming (SWJ), comb spectrum jamming (SPJ), and specific composite jamming (ISRJ+SWJ), the cluster shift magnitudes induced by both MI-FGSM and SSTA are relatively constrained. From the perspective of intrinsic signal analysis, these highly sparse spectrum or multi-domain strongly coupled jamming systems form extremely stable and isolated feature representation spaces in heterogeneous models. Under the strict infinity norm minimal perturbation constraint ($\epsilon=0.05$), the limited additive perturbation energy struggles to cross their broad decision boundaries. This not only accurately depicts the efficiency bottleneck of existing minimal additive adversarial frameworks in coping with highly robust physically isolated features but also points out a clear evolutionary direction for exploring non-additive or large-scale structural adversarial attacks in the future.

\subsubsection{Cross-Model Gradient Alignment and Optimization Dynamics}

\begin{figure*}[htbp]
    \centering
    \includegraphics[width=\textwidth]{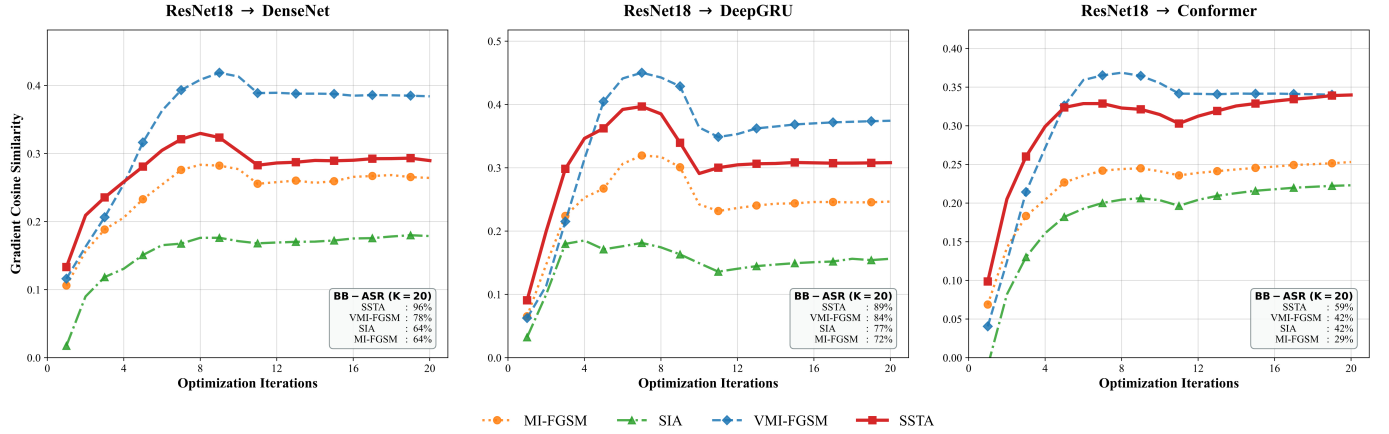}
    \caption{Dynamic evolution trajectories of cross-model gradient cosine similarity during the iteration process.}
    \label{fig:gradient}
\end{figure*}

The theoretical upper bound of the adversarial sample transfer success rate depends on the extent to which the optimization direction on the white-box source model can approximate the true descent direction of the black-box target model. To quantify the compatibility of this optimization dynamics, this paper introduces the Gradient Cosine Similarity metric across models. In specific implementation, to strictly eliminate the gradient noise interference caused by initial classification errors of the models, this experiment directionally screens 32 pure samples that can be correctly classified by the source model with $100\%$ confidence as the optimization base. Based on this standard subset, Fig. \ref{fig:gradient} tracks the real-time alignment trajectories of the four core algorithms during a 20-step iteration cycle. This experiment uses ResNet18 as the unified benchmark source model to calculate the similarity between its iterative surrogate gradients and the true gradients of three black-box models: DenseNet, DeepGRU, and Conformer.

Analyzing the dynamic evolution curves in Fig. \ref{fig:gradient}, in the early iteration stage ($K < 8$), the gradient alignment degree of all algorithms exhibits a rapid upward trend. This indicates that in the initial stage of perturbation exploration, all algorithms can effectively capture the shallow decision boundaries shared by the source and target models. However, as the iterations deepen, the gradient similarities of the four algorithms all experience significant degradation. This is because, in the middle and late stages, the perturbations become deeply trapped in the high-frequency Non-robust Features specific to the source model, causing their update directions to gradually deviate from the target model's true loss descent path, triggering feature-level ``source-model overfitting.'' Around $K=11$, constrained by the hard boundary of the $L_\infty$ norm truncation and the saturation effect of momentum accumulation, the exploration space of the adversarial perturbations peaks, and all four evolution curves gradually enter a dynamic plateau phase.

Throughout the entire plateau phase, the gradient cosine similarity of VMI-FGSM almost consistently overrides that of SSTA. However, cross-validating this with the terminal Black-Box Attack Success Rate (BB-ASR) at $K=20$ (shown in the bottom right corner of Fig. \ref{fig:gradient}), this experiment reveals a profound ``Optimization Paradox'': although VMI-FGSM achieves optimal mathematical gradient alignment in the surrogate space, SSTA comprehensively surpasses it in actual black-box transfer rates. Specifically, on the $\mathrm{ResNet18} \rightarrow \mathrm{DenseNet}$ link, SSTA significantly leads VMI-FGSM with an absolute success rate of $96\%$ to $78\%$; while on the most difficult link crossing to the global self-attention architecture ($\mathrm{ResNet18} \rightarrow \mathrm{Conformer}$), SSTA firmly holds a tactical deterrence baseline of $59\%$, far exceeding VMI-FGSM's $42\%$.

Analyzing this anomalous phenomenon from the underlying mechanisms of optimization theory and the Loss Landscape precisely reveals the physical root of SSTA's high transfer generalization capability. The variance tuning of VMI-FGSM is entirely confined to the static digital domain; its extremely high similarity means that the adversarial perturbations, at specific ``static input coordinates,'' precisely fall into the steep Sharp Minima basins where the source and target models coincidentally overlap. This stringent mathematical alignment severely lacks generalization tolerance and is extremely fragile when facing the architectural deformations of heterogeneous networks. In contrast, SSTA embeds actual time delays and amplitude-phase physical distortions as joint operators into the forward propagation, and its momentum gradient is the Expectation over Transformation (EoT) calculated under complex physical transformation distributions. This mechanism is equivalent to applying an Implicit Gradient Regularization based on physical priors to the loss extreme surface of the source model. Therefore, SSTA strategically sacrifices a portion of the ``static absolute alignment precision'' in exchange for the optimization trajectory evolving toward Flat Minima, which are smoother, broader, and possess exceptional structural robustness. It is exactly this smoothed solution, found under physical manifold constraints, that dynamically guarantees the extreme survivability of adversarial perturbations when penetrating unknown orthogonal defense architectures from the ground up.

\subsection{Physical Robustness and Effectiveness in EW Environments}

In the aforementioned digital-domain evaluations, the SSTA algorithm has demonstrated outstanding generalization capability across heterogeneous cognitive models. However, in highly dynamic, real-world radar electronic warfare environments, a purely digital-domain adversarial attack does not constitute a complete tactical closed loop. Reconnaissance and jamming signals inevitably experience receiver clock drift, electromagnetic channel fading, and hard-core coherent processing at the radar front-end. To comprehensively verify the practical deployment potential of adversarial examples, this section shifts the evaluation perspective from tensor space back to physical electromagnetic space, deeply demonstrating the immunity and robustness of the SSTA algorithm against physical distortions and its tactical equivalence in the radar coherent processing chain.

\subsubsection{Effective ASR under Physical Environment Distortions}

Traditional Attack Success Rate (ASR) cannot distinguish whether a model's prediction error is caused by the adversarial perturbation or solely by the physical environment distortion (e.g., severe noise blinding). To precisely quantify the true survival rate of adversarial perturbations under physical distortions, the Effective ASR metric is introduced. This metric utilizes a masking mechanism to calculate the subset of clean samples that the model still correctly predicts after noise injection at a specific physical distortion strength, and statistically evaluates the successful deception proportion of noisy adversarial samples only within this valid subset. Its rigorous mathematical definition is as follows:
\begin{equation} \label{eq:effective_asr}
\mathrm{Effective\_ASR} = \frac{\sum \mathbb{I}\left(F(\mathbf{x}_{\mathrm{clean}} + \mathbf{n}) = y \land F(\mathbf{x}_{\mathrm{adv}} + \mathbf{n}) \neq y\right)}{\sum \mathbb{I}\left(F(\mathbf{x}_{\mathrm{clean}} + \mathbf{n}) = y\right)}
\end{equation}
where $\mathbf{x}_{\mathrm{clean}}$ is the clean baseband signal, $\mathbf{x}_{\mathrm{adv}}$ is the adversarial example, $\mathbf{n}$ denotes the specific injected physical environment distortion, $y$ is the ground-truth label, $F(\cdot)$ represents the target cognitive model, and $\mathbb{I}(\cdot)$ is the indicator function.

Based on this evaluation framework, this paper tested the robustness degradation curves of the core baselines and the SSTA algorithm on a unified source model (ResNet18) under three extreme physical distortions: clock drift ($\Delta t$), static phase rotation ($\Delta \phi$), and Additive White Gaussian Noise channels (AWGN). Regarding the experimental setup, this phase first strictly screened and generated 256 test bases with complete adversarial features in the digital domain. Subsequently, in the physical distortion simulation, the time offset range was set to $[-30, 30]$ sampling points, the phase rotation range to $[-\pi, \pi]$ radians, and the Jamming-to-Noise Ratio (JNR) degradation range from $35\ \mathrm{dB}$ to $5\ \mathrm{dB}$. Each distortion dimension was uniformly divided into 31 fine-grained testing nodes to completely capture the performance evolution trajectory.

\begin{figure*}[htbp]
    \centering
    \includegraphics[width=\textwidth]{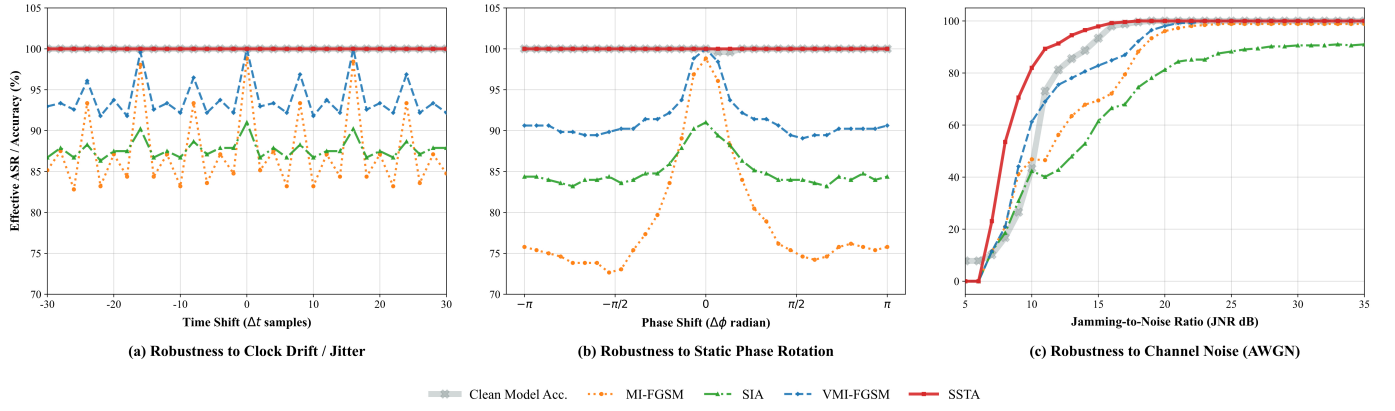}
    \caption{Effective ASR degradation curves under extreme physical distortions: clock drift, phase rotation, and AWGN channels.}
    \label{fig:robustness_degradation}
\end{figure*}

In the dynamic offset evaluation of the time and phase domains, constrained by the limitations of pure digital-domain optimization, the adversarial features of traditional algorithms are highly sensitive to absolute spatio-temporal coordinates. As shown in the distortion degradation curves in Fig. \ref{fig:robustness_degradation}, when time shifts ranging from $[-30, 30]$ sampling points or phase rotations of $[-\pi, \pi]$ are introduced, the Effective ASR of the baseline algorithms exhibits severe oscillations and cliff-like degradation, dropping to approximately $72\%$ in some intervals. Conversely, SSTA's evolution trajectory presents a perfect horizontal straight line across the entire spatio-temporal distortion interval, with its Effective ASR consistently anchored firmly at $100\%$. This proves from the underlying mechanism that the spatio-temporal shift operators introduced based on Expectation over Transformation (EoT) successfully endow the adversarial perturbations with extremely strong translation and phase rotation invariance.

In the more stringent AWGN channel evaluation, as the JNR decreases, the target model's recognition accuracy for clean samples (i.e., the theoretical upper-bound trajectory) inevitably declines accordingly. Observing the quantitative curves reveals that the perturbation energy generated by baseline algorithms rapidly collapses under the coverage of electromagnetic noise, and the decay rate of their effective success rate is much faster than the drop in the clean model's accuracy. In contrast, SSTA's performance curve presents a highly tactically valuable ``upper-bound adherence'' phenomenon: its evolution trajectory tightly and completely adheres to the theoretical upper-bound trajectory of the clean model's accuracy down to extremely low SNR intervals. This highly overlapping decay dynamic characteristic declares that until the physical channel noise completely blinds the defense model, SSTA's adversarial features will always maintain outstanding structural survivability.

\subsubsection{Tactical Equivalence Validation in Coherent Processing Chains}

Beyond resisting channel fading, adversarial perturbations must pass through the enemy radar's baseband coherent processing (such as matched filtering) to ensure the target generation capability of the original active deception jamming, that is, to maintain ``tactical equivalence.'' This paper takes typical Dense False Target Jamming (DFTJ) as an example to conduct high-fidelity demonstration of physical waveforms from the core dimensions of joint time-frequency distribution and pulse compression.

\begin{figure}[htbp]
    \centering
    \includegraphics[width=\columnwidth]{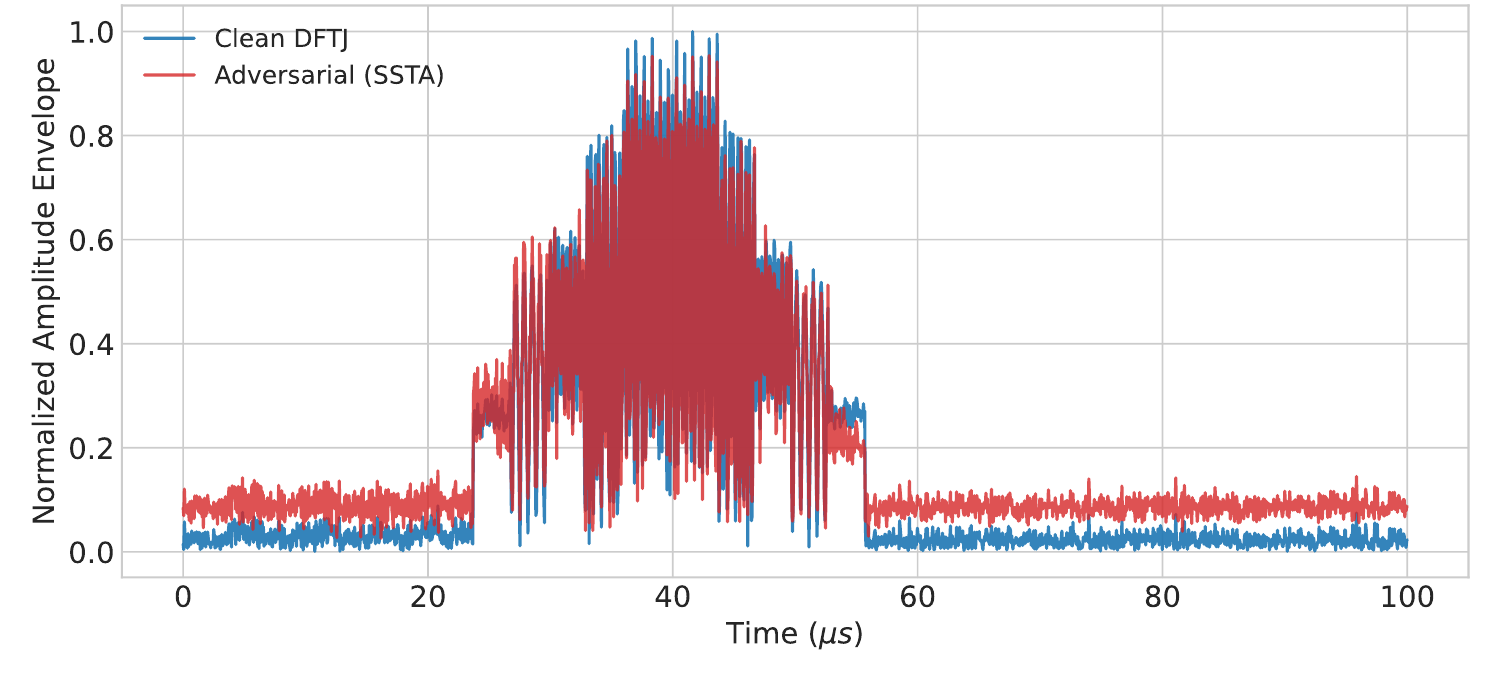}
    \caption{Normalized time-domain envelope comparison between the clean DFTJ signal and the SSTA adversarial example.}
    \label{fig:time_domain_envelope}
\end{figure}

\begin{figure}[htbp]
    \centering
    \includegraphics[width=\columnwidth]{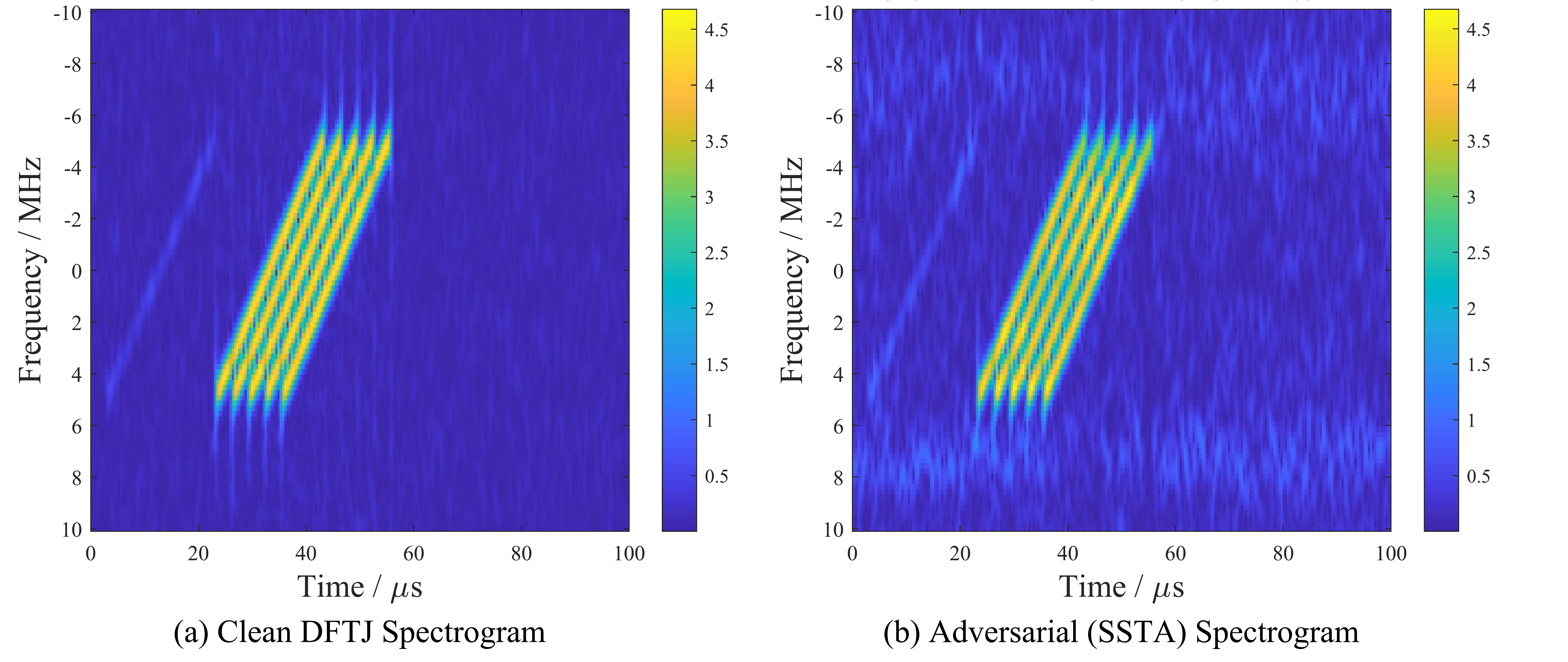}
    \caption{STFT spectrogram comparison: (a) Clean DFTJ signal; (b) SSTA adversarial example.}
    \label{fig:spectrogram_comparison}
\end{figure}

In the time-domain envelope (Fig. \ref{fig:time_domain_envelope}) and Short-Time Fourier Transform (STFT) spectrogram (Fig. \ref{fig:spectrogram_comparison}), benefiting from the strict minimal perturbation constraint ($\epsilon=0.05$), the physical distortions injected by SSTA are precisely limited within a minimal energy tolerance range. The adversarial examples do not cause macroscopic destruction to the temporal duty cycle of the original pulses, and the core false target energy ridges perfectly maintain the original linear frequency modulation slopes and delay intervals. The adversarial energy primarily diffuses into the background spectrum in a form similar to pseudo-random noise. This successfully misleads the cognitive network at the feature extraction level while presenting extremely high physical transparency to traditional radar time-frequency reconnaissance algorithms.

\begin{figure}[htbp]
    \centering
    \includegraphics[width=\columnwidth]{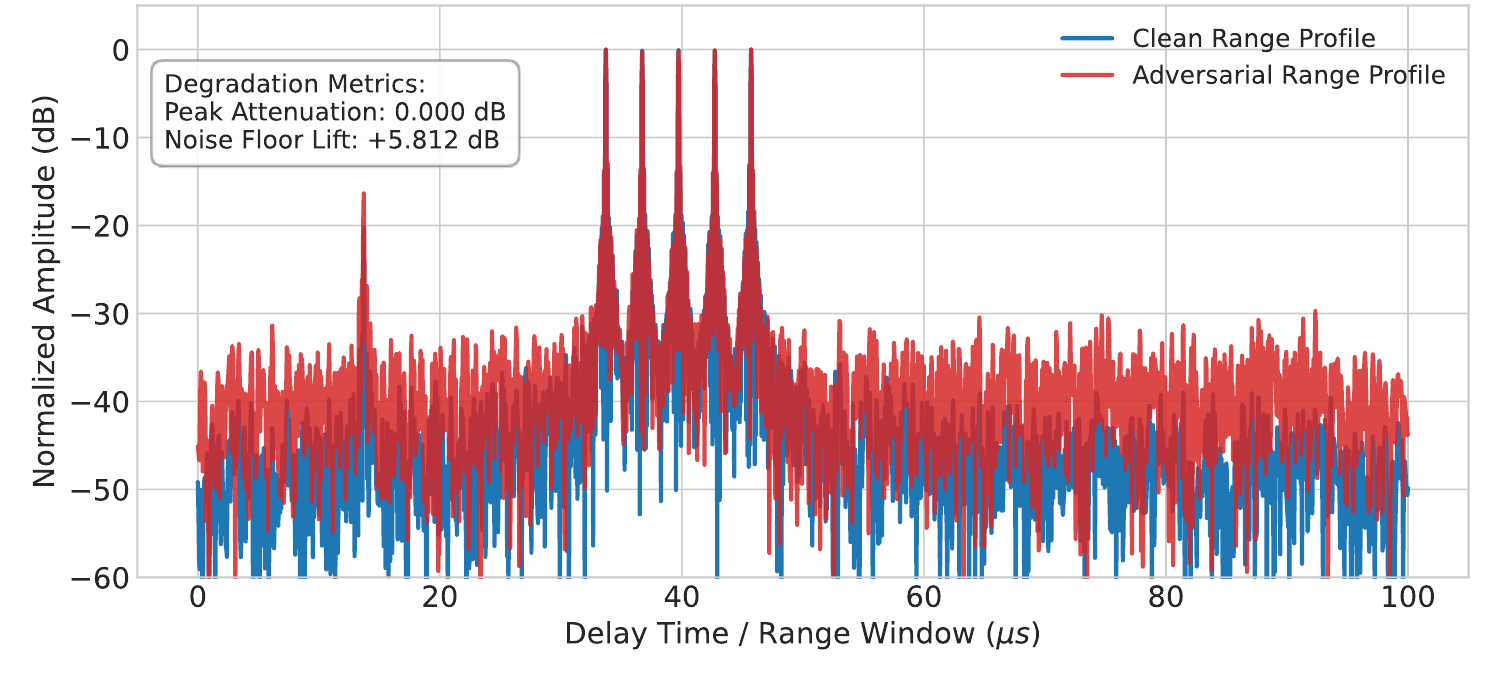}
    \caption{One-dimensional range profile comparison after matched filtering (pulse compression).}
    \label{fig:pulse_compression}
\end{figure}

In the pulse compression stage, which determines the radar's high-resolution range extraction, the one-dimensional Range Profile quantification metrics after matched filtering provide decisive physical evidence. Measured signal processing data reveals that the Peak Attenuation of the SSTA adversarial example is an absolute $0.000\ \mathrm{dB}$, meaning that SSTA induces zero loss to the core main-lobe energy of the original deception waveform. Furthermore, the Noise Floor Lift after matched filtering is marginally increased by only $+5.812\ \mathrm{dB}$. In practical radar system detection theory, this minor noise floor elevation is completely within the system tolerance closed-loop of the Constant False Alarm Rate (CFAR) detection mechanism; it will by no means mask the main-lobe energy of the dense false targets, nor will it trigger unreasonable false alarm thresholds. This dual characteristic of ``strong digital-domain blinding and physical-domain zero attenuation'' fundamentally establishes the extremely high application value of the SSTA algorithm in real electromagnetic countermeasure scenarios.

\section{Conclusion and Future Perspectives} \label{sec5}

With the evolution of cognitive radar systems toward the end-to-end processing of one-dimensional complex baseband sequences, the underlying adversarial security vulnerabilities of deep networks have become increasingly prominent. Addressing the dual limitations of existing adversarial attack methods in cross-architecture transferability and physical receiving chain compatibility, this paper proposes a physical prior-driven Segmented Spatio-Temporal Attack (SSTA). By employing a segmentation mechanism based on global-local feature decoupling, SSTA effectively blocks the dependence of adversarial gradients on the long-range coherence of a single network, mitigating the problem of traditional adversarial perturbations overfitting to source models. In cross-evaluations involving 9 heterogeneous models spanning CNNs, RNNs, and Transformers, the Average Black-Box Attack Success Rate (ABB-ASR) of SSTA is improved by over 10\% compared to existing baseline algorithms, demonstrating the effectiveness of discrete local optimization in extracting highly generalizable adversarial features. Furthermore, by constructing spatio-temporal physical operators combined with the Expectation over Transformation (EoT) strategy, SSTA achieves an effective integration of adversarial iterations with actual channel distortions (delay and amplitude-phase fluctuations). Radar baseband equivalent simulation tests indicate that, after penetrating the radar matched filter, the generated adversarial waveforms cause no obvious attenuation to the main-lobe peak ($0.000\ \mathrm{dB}$), and the noise floor lift is strictly controlled within the Constant False Alarm Rate (CFAR) detection tolerance. This implies that while inducing high-confidence misclassifications in cognitive models, SSTA effectively preserves the coherent envelope structure of the original deception jamming, achieving the tactical unification of digital-domain deception and physical-domain fidelity.

Although the proposed SSTA framework effectively reveals the adversarial vulnerabilities faced by 1D radar baseband cognitive networks, certain limitations remain when transitioning to practical engineering deployment. On the one hand, the joint optimization mechanism based on expectation over transformation and multi-step momentum iterations introduces substantial computational overhead, making it difficult to fully satisfy the microsecond-level real-time response constraints in highly dynamic electronic warfare. On the other hand, the currently constructed spatio-temporal physical operators primarily focus on direct-path delays and basic amplitude-phase distortions, lacking comprehensive coverage of higher-order environmental effects induced by high-speed platform maneuvers and complex terrains, such as Doppler dispersion and multipath fading. To address these shortcomings, future research will focus on two core dimensions: ``lightweight real-time processing'' and ``physical closed-loop validation.'' At the algorithmic level, we plan to explore forward inference networks based on deep generative architectures to replace the time-consuming iterative optimization process, aiming to achieve single-step real-time generation of adversarial waveforms and incorporate more comprehensive dynamic channel prior models. At the physical validation level, future work will be dedicated to deploying the adversarial generation framework onto real radio frequency hardware platforms, such as Software-Defined Radio (SDR), to strictly calibrate the RF survivability of adversarial examples in open Over-the-Air (OTA) channels. Additionally, extending the existing 1D temporal adversarial mechanisms to distributed cognitive radar networks, and investigating the blinding evolution laws of adversarial waveforms in multi-static collaborative detection and data fusion centers, will also serve as vital exploratory directions for deepening the security assessment of cognitive electronic warfare.

\bibliographystyle{IEEEtran}
\bibliography{ref}
\newpage
\vfill
\end{document}